\def\year{2022}\relax
\documentclass[letterpaper]{article} % DO NOT CHANGE THIS
\usepackage{aaai22}  % DO NOT CHANGE THIS
\usepackage{times}  % DO NOT CHANGE THIS
\usepackage{helvet}  % DO NOT CHANGE THIS
\usepackage{courier}  % DO NOT CHANGE THIS
\usepackage[hyphens]{url}  % DO NOT CHANGE THIS
\usepackage{graphicx} % DO NOT CHANGE THIS
\usepackage{natbib}  % DO NOT CHANGE THIS AND DO NOT ADD ANY OPTIONS TO IT
\usepackage{caption} % DO NOT CHANGE THIS AND DO NOT ADD ANY OPTIONS TO IT
\DeclareCaptionStyle{ruled}{labelfont=normalfont,labelsep=colon,strut=off} % DO NOT CHANGE THIS
\usepackage{algorithm}
\usepackage{algorithmic}

\usepackage{hyperref}       % hyperlinks: cannot be used in AAAI format
\usepackage{url}            % simple URL typesetting
\usepackage{booktabs}       % professional-quality tables
\usepackage{amsfonts}       % blackboard math symbols
\usepackage{nicefrac}       % compact symbols for 1/2, etc.
\usepackage{microtype}      % microtypography
\usepackage{xcolor}         % colors

\usepackage{enumitem} % Control layout of itemize, enumerate, description
\usepackage{amsmath}	% AMS mathematical facilities for LATEX

\usepackage{graphicx}
\usepackage{subfigure}
\usepackage{booktabs} % for professional tables
\usepackage[export]{adjustbox}

\usepackage{comment}

\usepackage{newfloat}
\usepackage{listings}
\floatstyle{ruled}
\newfloat{listing}{tb}{lst}{}
\floatname{listing}{Listing}
\title{ Blockwise Sequential Model Learning for \\ Partially Observable Reinforcement Learning }

\author {
    Giseung Park, Sungho Choi, Youngchul Sung
}
\affiliations {
  School of Electrical Engineering, KAIST, Korea \\
    \{gs.park, sungho.choi, ycsung\}@kaist.ac.kr
}

\begin{document}

\maketitle

\begin{abstract}
This paper proposes a new sequential model learning architecture to solve partially observable Markov decision problems.  Rather than compressing sequential information at every timestep as in conventional recurrent neural network-based methods, the proposed architecture generates a latent variable in each data block with a length of multiple timesteps and passes the most relevant information to the next block for policy optimization. The proposed blockwise sequential model is implemented based on self-attention, making the model capable of detailed sequential learning in partial observable settings. The proposed model builds an additional learning network to efficiently implement gradient estimation by using self-normalized importance sampling, which does not require the complex blockwise input data reconstruction in the model learning. Numerical results show that the proposed method significantly outperforms previous methods in various partially observable environments.
\end{abstract}

%%%%%%%%%%%%%%%%%%%%%%%%%%%%%%%%%%%%%%
\section{Introduction}
%%%%%%%%%%%%%%%%%%%%%%%%%%%%%%%%%%%%%%

Reinforcement learning (RL) in partially observable environments is usually formulated as partially observable Markov decision processes (POMDPs). RL solving POMDPs is a challenging problem since the Markovian assumption on observation is broken. The information from the past should be extracted and exploited during the learning phase to compensate for the information loss due to partial observability. Partially observable situations are prevalent in real-world problems such as control tasks when observations are noisy, some part of the underlying state information is deleted, or long-term information needs to be estimated \cite{han20vrm,meng21memory}. 

Although many RL algorithms have been devised and state-of-the-art algorithms provide outstanding performance in fully observable environments, relatively fewer methods have been proposed to solve POMDPs. Previous POMDP methods use a recurrent neural network (RNN) either to compress the information from the past in a model-free manner \cite{hausknecht15drqn,zhu17adrqn,goyal21rims} or to estimate the underlying state information and use the estimation result as an input to the RL agent \cite{igl18dvrl,han20vrm}. These methods compress observations in a  step-by-step sequential order in time, which may be inefficient when partiality in observation is high and less effective in extracting {\em contextual} information within a time interval.

\begin{figure}[!t] %%% Figure 1. Proposed overall architecture
	\begin{center} 
		\includegraphics[width=\linewidth]{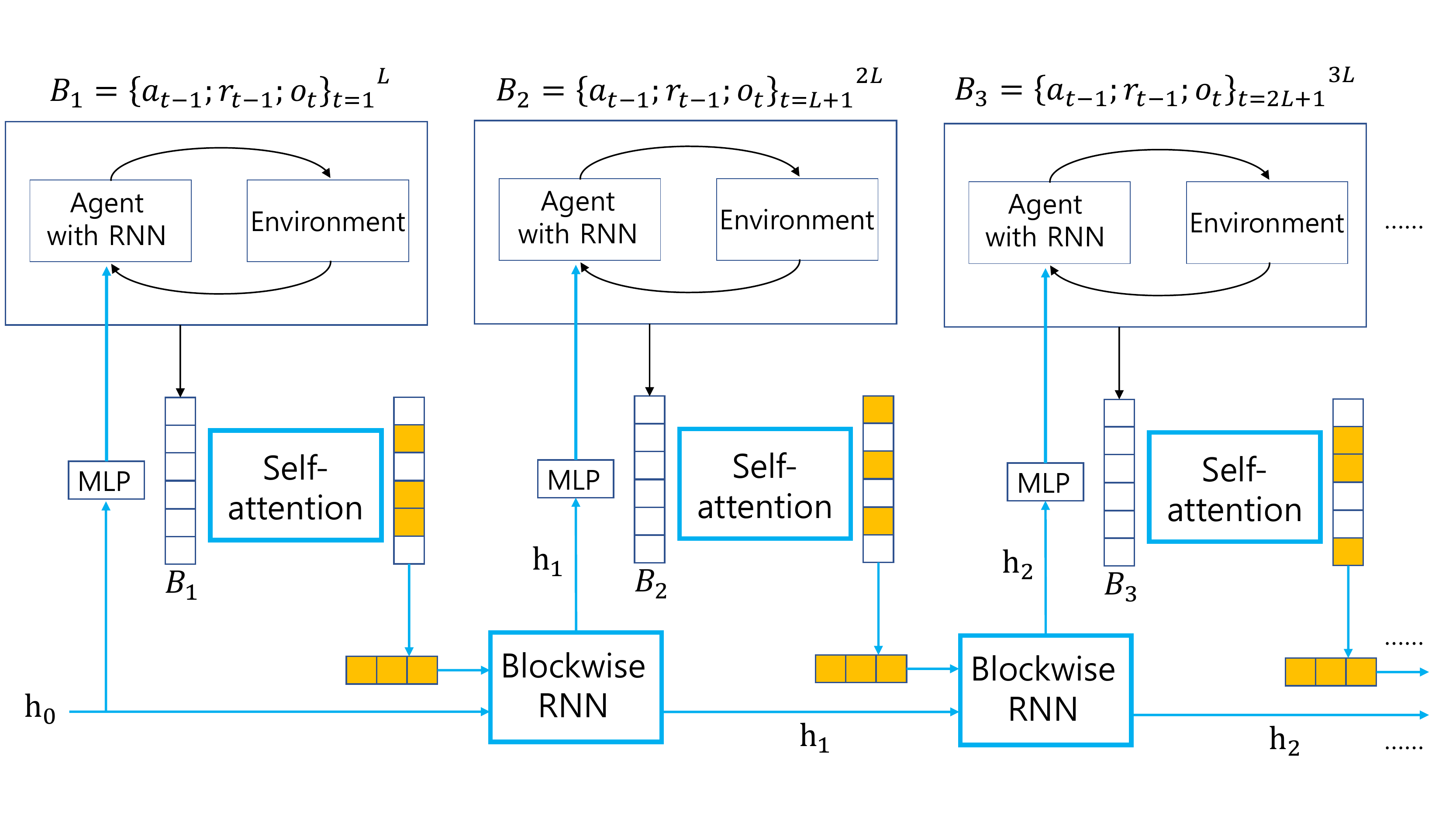}
		\caption{  Overall architecture of the proposed model: Self-attention and RNN are used to process each block $B_n$ of sequential data instead of processing at every timestep.  }
		\label{fig:high_level_proposed_architecture}
	\end{center}
% 	\vspace{-0.2cm}
\end{figure}

We conjecture that observations at specific timesteps in a given time interval contain more information about decision-making. We propose a new architecture to solve partially observable RL problems by formalizing this intuition into a mathematical framework. Our contributions are as follows:

\begin{itemize}[leftmargin=5mm]
    
    \item As shown in Fig. \ref{fig:high_level_proposed_architecture}, we propose a new learning architecture based on  a \textbf{block} of sequential input rather than estimating a latent variable at every timestep by jointly using self-attention \cite{vaswani17attention}  and RNN and 
exploiting the advantage of each structure.
    
    \item To learn the proposed architecture, we present a blockwise sequential model learning based on direct gradient estimation  using self-normalized importance sampling \cite{bornschein14rws,le19revisitrws}, which does not require input data reconstruction in contrast to usual variational methods to POMDPs \cite{chung15vrnn,han20vrm}. 
    
    \item Using the proposed blockwise representations of the proposed model and feeding the learned block variables to the RL agent, we significantly improved the performance over existing methods in several POMDP environments.

\end{itemize}

%%%%%%%%%%%%%%%%%%%%%%%%%%%%%%%%%%%%%%%%%
\section{Related Work}
%%%%%%%%%%%%%%%%%%%%%%%%%%%%%%%%%%%%%%%%%

In partially observable RL, past information should be exploited appropriately to compensate for the information loss in the partial observation.  RNN and its variants \cite{hochreiter97lstm,cho14gru} have been  used to process the past information. The simplest way is that the output of RNN driven by the sample sequence  is directly fed into the RL agent as the input capturing the past information without further processing, as considered in previous works \cite{hausknecht15drqn,zhu17adrqn}. The main drawback of these end-to-end approaches is that it requires considerable data for training RNN and is suboptimal in some complicated environments \cite{igl18dvrl,han20vrm}.

\citet{goyal21rims} proposed a variant of RNN in which  the hidden variable is divided into multiple segments with equal length. First, a fixed number of the segments are selected using attention \cite{vaswani17attention}. Then, only the selected segments are updated with independent RNNs followed by self-attention, and the remaining segments are not changed. Our approach is substantially different from this method in that we use attention over a time interval, while the structure of \citet{goyal21rims} is updated stepwise by using the attention over the segments at the same timestep.

Other methods estimate state information or belief state by learning a sequential model of stepwise latent variables. The inferred latent variables are then used as input to the RL agent.  \citet{igl18dvrl} proposed estimating the belief state by applying a particle filter \cite{maddison17fivo,le18aesmc,naesseth18vsmc} in variational learning. \citet{han20vrm} proposed a Soft Actor-Critic \cite{haarnoja18sac} based method (VRM) focusing on solving partially observable continuous action control tasks.  VRM adds action sequence as additional input and uses samples from the replay buffer to maximize the variational lower bound \cite{chung15vrnn}. Then, latent variables are generated as input to the RL agent. To solve the stability issue, however, VRM concatenates (i) a pre-trained and frozen variable $d_{\text{freeze}}$, and (ii) a learned variable $d_{\text{keep}}$ from the distinct model because using only $d_{\text{keep}}$ as  input to  the RL agent does not yield performance improvement. In contrast, our method only uses one block model for learning, which is more efficient.

While previous methods improved performance in partially observable environments, they mostly use  RNN. The RNN architecture suffers two problems when partiality in observation is high: (i) the forgetting problem and (ii) the inefficiency of stepwise compressing all the past samples, including unnecessary information such as noise. Our work solves these problems by learning our model blockwise by passing the most relevant information to the next block.

\section{ Background }
\label{section:background}

\begin{figure*}[!ht] %%% Figure 1. Proposed overall architecture of ARMA
	\begin{center} 
		\includegraphics[width=0.77\linewidth]{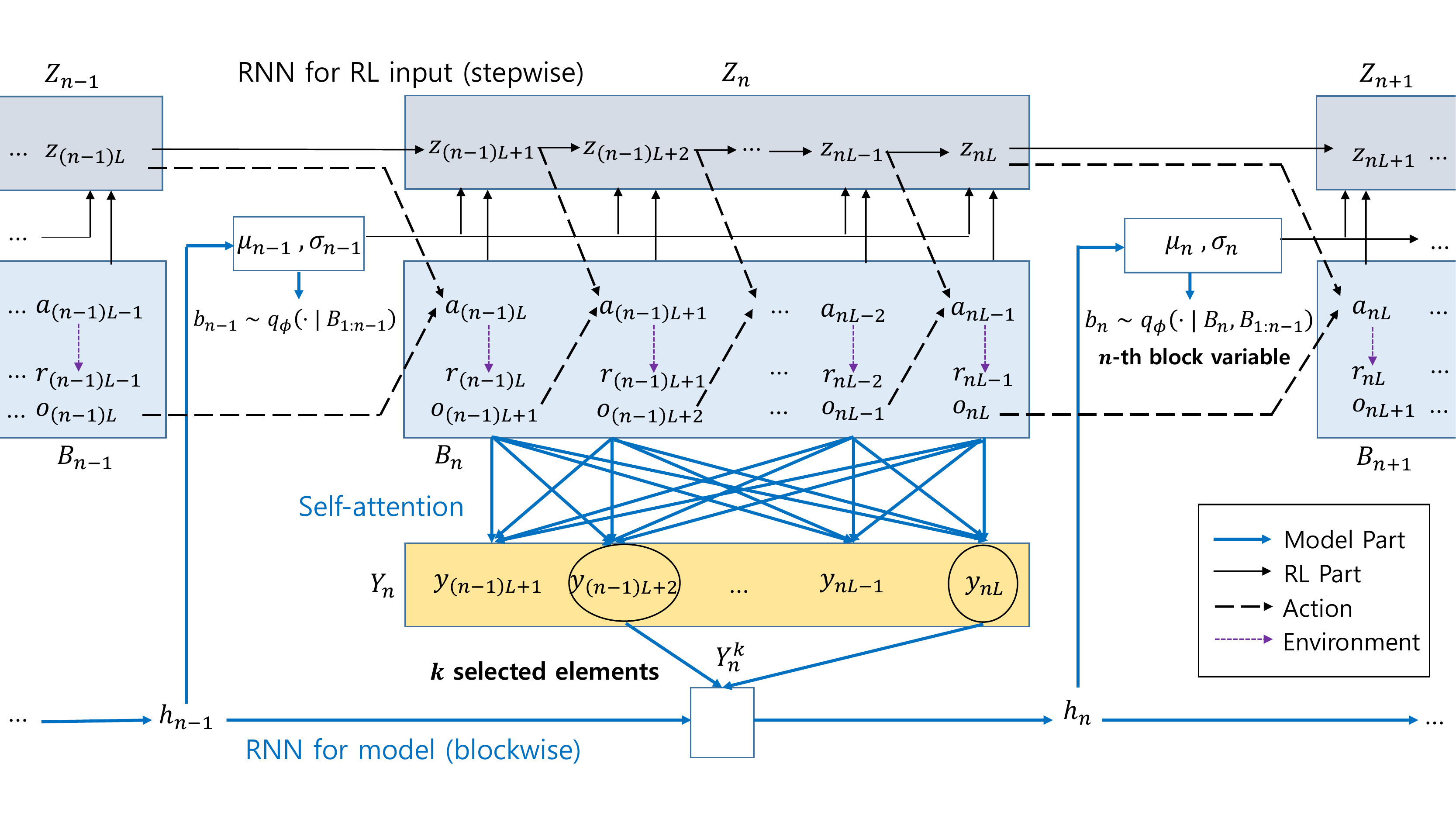}
		\vspace{-6mm}
		\caption{ Detailed architecture of the proposed model: An analogy to filter theory can be drawn. RNN corresponds to autoregressive (AR) filtering, which performs recursive filtering only, attention corresponds to moving-average (MA) filtering, which performs block processing, and the proposed new architecture corresponds to autoregressive--moving-average (ARMA) filtering. }
		\label{fig:proposed_architecture}
	\end{center}
	\vspace{-2mm}
\end{figure*}

\subsubsection{Setup}

We consider a discrete-time  POMDP denoted by  $(\mathcal{S}, \mathcal{A}, P, r, \gamma, \mathcal{O}, \Omega)$, where   $\mathcal{S}$ and $\mathcal{A}$ are the state and action spaces,  respectively, $P: \mathcal{S} \times \mathcal{A} \times \mathcal{S} \rightarrow \mathbb{R}^+$ is the state transition probability distribution,   $r: \mathcal{S} \times \mathcal{A} \times \mathcal{S}  \rightarrow \mathbb{R}$ is the reward function, $\gamma \in [0, 1]$ is the discounting factor, and $\mathcal{O}$ and $\Omega$ are the observation space and observation probability, respectively.  Unlike in a usual MDP setting, the agent cannot observe the state $s_t$ at timestep $t$ in POMDP, but receives an observation $o_t \in \mathcal{O}$ which is generated by the observation probability $\Omega: \mathcal{S} \times \mathcal{A} \times \mathcal{O} \rightarrow \mathbb{R}^+$.  Our goal is to optimize  policy $\pi$ to maximize the expected discounted  return $\mathbb{E}_{ \pi}[\sum_{t=0}^\infty \gamma^t r_{t}]$ by learning $\pi$ with  a properly designed input variable to $\pi$ in addition to $o_t$  in place of the unknown true state $s_t$  at each timestep $t$.

\subsubsection{Self-attention}

Self-attention \cite{vaswani17attention} is an architecture that can perform a detailed process within a time interval by considering contextual information among sequential input data in the interval.  Consider a sequential input data of length $L$, denoted by $B = x_{1:L} \stackrel{\triangle}{=} [x_1, x_2, \cdots, x_L]^{\top} \in \mathbb{R}^{L \times d}$, where  $x_i \in \mathbb{R}^d$ (column vector), $1 \leq i \leq L$, and $(\cdot)^\top$ denotes matrix transpose. (The notation $A_{m_1:m_2}\stackrel{\triangle}{=} [A_{m_1},A_{m_1+1},\cdots,A_{m_2}]$ for any quantity $A$ will be used in the rest of the paper.)
Self-attention architecture transforms each input data $x_i$ in $B$ into $y_i$ so that the transformed representation $y_i$ contains information in not only $x_i$ but also all other $x_j \in B$, reflecting the relevance to the target task. (See Appendix \ref{sec:append_self_attention} for the structure.)
% A: \ref{sec:append_self_attention}

To improve the robustness of  learning, self-attention is usually implemented with $m~(>1)$ multi-head transformation.  Let the $d \times d$ transform matrices of query, key, and value be $M^Q = [M_1^Q, M_2^Q, \cdots, M_m^Q], ~ M^K = [M_1^K, M_2^K, \cdots, M_m^K], ~ M^V = [M_1^V, M_2^V, \cdots, M_m^V]$, respectively, where $d = m h_{\text{head}}$ so that $M_l^Q, M_l^K, M_l^V \in \mathbb{R}^{d \times h_{\text{head}} } $ for each $1 \leq l \leq m$. The $l$-th query, key, and value are defined as $B M_l^Q, B M_l^K, B M_l^V \in \mathbb{R}^{L \times h_{\text{head}} } $, respectively. Using an additional transform matrix $M^O \in \mathbb{R}^{ m h_{\text{head}} \times d } = \mathbb{R}^{ d \times d } $, the output of multi-head self-attention $\text{MHA}(B) \in \mathbb{R}^{L \times d }$ is given by
\begin{align} \label{eq:multihead}
\text{MHA}(B) &=  [A_1, A_2, \cdots, A_m]M^O, ~~\mbox{where}~~ \nonumber \\
A_l &= f \left( \frac{(B M_l^Q) (B M_l^K)^\top }{\sqrt{h_{\text{head}}}} \right) (B M_l^V) \in \mathbb{R}^{L \times h_{\text{head}} }   
\end{align}
and  $f$ is a row-wise softmax function (other pooling methods can be used in some cases \cite{richter20normattention}).

In practice, residual connection, layer normalization \cite{ba16layernorm}, and a feed-forward neural network $g$ are used to produce the final $L \times d$ representation $Y = [y_1, y_2, \cdots, y_L]^\top$:
\begin{align} \label{eq:attention_all}
Y &= \text{LayerNormalize}(g(U) + U), ~~\mbox{where}~~ \nonumber \\
U &= \text{LayerNormalize}(\text{MHA}(B) + B).  
\end{align}
Note that the self-attention architecture of $B \rightarrow Y$ can further be stacked multiple times for deeper representation.

Unlike RNN, however, in self-attention, each data block is processed without consideration of the previous  blocks, and hence each transformed block data is disconnected. Therefore, information from the past is not used to process the current block. In contrast, RNN uses past information by stepwise accumulation, but RNN suffers the forgetting problem when the data sequence becomes long.

%%%%%%%%%%%%%%%%%%%%%%%%%%%%%%%%%%%%%
\section{Proposed Method} \label{sec:proposed_method}
%%%%%%%%%%%%%%%%%%%%%%%%%%%%%%%%%%%%%

We present a new architecture for POMDPs,  modeling blockwise latent variables by jointly using self-attention and RNN and exploiting the advantage of each structure.  The proposed architecture consists of (i) stepwise RNN for RL input and (ii) block model. If only the stepwise RNN is used, it corresponds to the naive RNN method. As shown in Figs. \ref{fig:high_level_proposed_architecture} and \ref{fig:proposed_architecture}, the  block model consists of self-attention and blockwise RNN. After the self-attention compresses block information, the blockwise RNN passes the information to the next block.

We describe the block model structure in Section \ref{subsec:PropArch}, including how the block model is used for the trajectory generation and how the block information is compressed. Section \ref{subsec:learning} explains how the block model is efficiently learned to help the RL update.

\subsection{Proposed Architecture}\label{subsec:PropArch}

We consider a sample sequence  $\{ x_t, t=1,\cdots,T  \}$  of length $T=NL$, where the sample $x_t \in \mathbb{R}^d$ at timestep $t$
is given by the column-wise concatenation $x_t = [a_{t-1}; r_{t-1}; o_{t}  ]$  of action $a_{t-1}$, reward $r_{t-1}$, and the partial observation $o_t$.  We partition the sample sequence $\{ x_t, t=1,\cdots,T  \}$  into $N$ blocks $B_1,\cdots,B_N$, where  the $n$-th block $B_n \in \mathbb{R}^{L \times d} ~ (1 \leq n \leq N)$  is given by 
\[
B_n = x_{(n-1)L+1:nL} = [x_{(n-1)L+1}, x_{(n-1)L+2}, \cdots, x_{nL}]^\top.
\]
We then decompose the model-learning objective as 
\begin{equation} \label{eq:log_objective}
    \log p_\theta (x_{1:T}) = \sum_{n=1}^{N} \log p_\theta ( B_n | B_{1:n-1} ),
\end{equation}
instead of conventional sample-by-sample decomposition
$\log p_\theta (x_{1:T}) = \sum_{t=1}^{T} \log p_\theta ( x_t | x_{1:t-1} )$.
Note that the conditioning term in \eqref{eq:log_objective} is $B_{1:n-1}$ not $B_{n-1}$   since the full Markovian assumption is broken in partially observable environments.  Based on this block-based decomposition, we define the generative model $p_\theta(B_{1:N}, b_{1:N})$ and the inference model $q_\phi( b_{1:N} | B_{1:N})$ as  
\begin{align} \label{eq:gen_and_inf_models}
 p_\theta(B_{1:N}, b_{1:N}) &= \prod_{n=1}^N p_\theta ( B_n, b_n | B_{1:n-1} ), \nonumber \\
 q_\phi( b_{1:N} | B_{1:N}) &= \prod_{n=1}^N q_\phi (  b_n |B_n, B_{1:n-1} ),
\end{align}
where $b_n$ is a latent variable containing the information of the $n$-th block $B_n$.

For each block index $n$, we want to infer the $n$-th block variable $b_n$ from the amortized posterior distribution $q_\phi$ in \eqref{eq:gen_and_inf_models} by using (i) the information of the current block $B_n$ and (ii) the information from past blocks $B_{1:n-1}$ before $B_n$. After inferring $b_n$ at the $n$-th block, the information of $b_n$ is used to generate the input variables $Z_n \stackrel{\triangle}{=} z_{(n-1)L+1:nL}$ to the RL agent. Then, the RL agent learns  policy $\pi$ based on $\{  (z_t, o_t, a_t, r_t), ~t=1,2,3,\cdots\}$, where $z_t$ is extracted from $Z_n$ containing $z_t$. Action $a_t$ is taken by the agent based on partial observation $o_t$ and additional input $z_t$ compensating for partiality in observation $o_t$ according to $a_t \sim  \pi(\cdot|z_t,o_t)$.

\subsubsection{ Trajectory Generation }
Fig. \ref{fig:proposed_architecture} shows the proposed architecture with  the $n$-th block  processing  as reference. 
The blue arrows in the lower part show the block variable inference network $q_\phi$, and the solid black arrows in the upper part represent the processing network for RL learning. 

%% Inner-block generation: z for RL part
Until  block index $n-1$ (i.e., timestep $t = (n-1)L$), the information from the previous blocks $B_{1:n-1}$ is compressed into the variable $h_{n-1}$. The $(n-1)$-th block latent variable $b_{n-1}$ is generated according to $b_{n-1} \sim q_{\phi} (\cdot | B_{1:n-1}) = \mathcal{N}(\mu_{n-1}, \text{diag} ( \sigma_{n-1}^2 )  )$, where  $\mu_{n-1}$ and $\sigma_{n-1}$ are the outputs of two neural networks with input $h_{n-1}$. The information of stochastic $b_{n-1}$ is summarized in $\mu_{n-1}$ and $\sigma_{n-1}$, so these two variables together with samples $B_n = x_{(n-1)L+1:nL}$ are fed into the RL input generation RNN to sequentially generate the RL input variables $Z_n = z_{(n-1)L+1 : nL}$ during the $n$-th block period. (Note that  $\mu_{n-1}$ and $\sigma_{n-1}$ capture the information in $h_{n-1}$.)  

The stepwise RL processing is as follows: Given the RL input $z_{(n-1)L}$ and the observation $o_{(n-1)L}$ from the last timestep $t=(n-1)L$ of $B_{n-1}$, the RL agent selects an action  $a_{(n-1)L} \sim \pi(\cdot | z_{(n-1)L}, o_{(n-1)L})$ (see the dashed black arrows). Then, the environment returns the reward $r_{(n-1)L}$ and the next observation $o_{(n-1)L + 1}$. Then, the sample $x_{(n-1)L + 1} = [a_{(n-1)L}; r_{(n-1)L}; o_{(n-1)L + 1}  ]$ at timestep $t=(n-1)L+1$  together with $\mu_{n-1}$ and $\sigma_{n-1}$ is  fed into the stepwise RNN  to produce the next RL input $z_{(n-1)L + 1}$. This execution is repeated at each timestep until $t= nL$ to produce $Z_n = z_{(n-1)L+1:nL}$, and each sample $x_t$ is stored in a current batch (on-policy) or a replay memory (off-policy).

% The latent variable generation capturing the contextual  information in each block is as follows. 
At the last timestep $t= nL$ of the $n$-th block, the $n$-th block data $B_n = x_{(n-1)L + 1 : nL}$ is fed into the {\em self-attention network} to produce the output $Y_n \stackrel{\triangle}{=} y_{(n-1)L + 1 : nL}$ capturing the contextual information in $B_n$. The procedure extracting $Y_n$ from $B_n$ follows the standard self-attention processing. However, instead of using all $Y_n$ to represent the information in $B_n$,  we  select   $k~(<L)$ elements in $Y_n$  for data compression and efficiency. We denote the concatenated vector of the $k$ elements by $Y_n^k$. $Y_n^k$ is fed into the blockwise RNN, which compresses  $Y_n^k$ together with $h_{n-1}$ to produce $h_n$. Thus, $h_n$ has the compressed information up to the $n$-th block $B_{1:n}$. The impact of the self-attention network and the blockwise RNN is analyzed in Section \ref{subsec:ablation_components_compare}.

\subsubsection{ Block Information Compression }
In order to select  $k$  elements  from $Y_n$, we  exploit the self-attention structure and the weighting matrix $W_l \stackrel{\triangle}{=} f \left( \frac{(B_n M_l^Q) (B_n M_l^K)^T }{\sqrt{h_{\text{head}}}} \right) (1 \leq l \leq m, m ~ \text{is the number of multi-heads})$ appearing in \eqref{eq:multihead}. The $p$-th column of the $j$-th row of $W_l$ determines the importance of the $p$-th row  of  $B_nM_l^V$ (i.e., data at timestep $(n-1)L+p$) to produce the $j$-th row of the attention $A_l$ in \eqref{eq:multihead}.  

Hence, adding all elements in the $p$-th column of whole matrix $W_l$, we can determine the overall contribution (or importance) of the $p$-th row  of  $B_nM_l^V$ to generate $A_l$. We choose the $k$ positions with largest $k$ contributions in column-wise summation of $\frac{1}{m} \sum_{l=1}^m W_l$ in timestep and choose the corresponding $k$ positions in $Y_n$ as our representative attention messages, considering the one-to-one mapping from $B_n$ to $Y_n$. The effect of the proposed compression method is analyzed in Section \ref{subsec:ablation_compression_compare}.
% $Y_n^k$

\subsection{ Efficient Block Model Learning } \label{subsec:learning}

The overall learning is composed of two parts: block model learning and RL policy learning. First, we describe block model learning. Basically, the block model learning is based on maximum likelihood estimation (MLE) to maximize the likelihood \eqref{eq:log_objective} for given data $x_{1:T}$ with the generative model $p_\theta$ and the inference model $q_\phi$ defined in \eqref{eq:gen_and_inf_models}.

In conventional variational approach cases such as variational RNN \cite{chung15vrnn},  the generative model $p_{\theta}$ is implemented as the product of a prior latent distribution and a decoder distribution, and the decoder is learned to reconstruct each input sample $x_t$ given the latent variable at each timestep $t$.  In our blockwise setting allowing attention processing, however, learning to reconstruct the block input $B_n = x_{(n-1)L + 1 : nL}$ with a decoder given a single block variable $b_n$ is challenging and complex  compared to the stepwise variational RNN case. 

In order to circumvent this difficulty, we approach the MLE problem based on {\em  self-normalized importance sampling} \cite{bornschein14rws,le19revisitrws}, which does not require an explicit reconstruction procedure. Instead of estimating the value of $p_\theta ( B_n | B_{1:n-1} )$, 
we  directly estimate the gradient $\nabla_{\theta} \log p_\theta ( B_n | B_{1:n-1} ) $ by using self-normalized importance sampling to update the generative model parameter $\theta$ (as $\theta \rightarrow \theta + c \nabla_{\theta} \log p_\theta ( B_n | B_{1:n-1} )$) to maximize the log-likelihood $\log p_\theta ( B_n | B_{1:n-1} ) $ in \eqref{eq:log_objective}.

% B: \ref{sec:append_grad_estimate}
The detailed procedure is as follows. (The overall learning procedure is detailed in Appendix \ref{sec:append_grad_estimate}.)  To estimate the gradient $\nabla_{\theta} \log p_\theta ( B_n | B_{1:n-1} ) $, we construct a  neural network parameterized by $\theta$ which produces the value of $\log p_\theta ( B_n, b_n |    B_{1:n-1}  )$. (The output of this neural network is  the logarithm of $p_\theta (B_n,b_n| B_{1:n-1})$ not $p_\theta (B_n,b_n| B_{1:n-1})$ for convenience.)  Using the formula  $ \nabla_{\theta} p_\theta ( B_n | B_{1:n-1} ) = \int \nabla_{\theta} p_\theta ( B_n, b_n | B_{1:n-1} ) d b_n $, we can express the gradient  $g_\theta^n := \nabla_{\theta} \log p_\theta ( B_n | B_{1:n-1} )$ as
\begin{equation} \label{eq:generative_learn}
     g_\theta^n \approx  \sum_{j=1}^{ K_{\text{sp}} } \frac{w_n^j}{  \sum_{j'=1}^{ K_{\text{sp}} } w_n^{j'}}   \nabla_{\theta} \log p_\theta ( B_n, b_n^j |    B_{1:n-1}  ),
\end{equation}
where  $w_n^j = \frac{  p_\theta ( B_n, b_n^j |    B_{1:n-1}  )   }{ q_\phi ( b_n^j | B_n, B_{1:n-1} )  }$ and $b_n^j  \sim   q_\phi ( b_n | B_n, B_{1:n-1} )$ for $1\leq j \leq K_{\text{sp}}$. (See Appendix \ref{sec:append_grad_estimate} for the full derivation.) The numerator of the importance sampling ratio $w_n^j$ can be computed as $\exp(\log p_\theta ( B_n, b_n^j |    B_{1:n-1}))$ based on the output of the constructed  generative model yielding $\log p_\theta ( B_n, b_n |    B_{1:n-1})$. The denominator $q_\phi ( b_n | B_n, B_{1:n-1} )$ is modeled as Gaussian distribution   $b_n \sim \mathcal{N}(\mu_{n}, \text{diag} ( \sigma_{n}^2 )  )$, where $\mu_n$ and $\sigma_n$ are functions of $h_n$. $h_n$ itself is a function of the blockwise RNN and the self-attention module, as seen in Fig. \ref{fig:proposed_architecture}. Thus, the parameters of the blockwise RNN and the self-attention module  in addition to the parameters of the $\mu_n$ and $\sigma_n$ neural network with input $h_n$ constitute the whole model parameter $\phi$.

Note that proper learning of $q_\phi$ is required to estimate $\nabla_{\theta} \log p_\theta ( B_n | B_{1:n-1} ) $ accurately in \eqref{eq:generative_learn}. For this, we learn $q_\phi$ to minimize 
\[
D_n \stackrel{\triangle}{=} D_{KL} [ p_\theta ( b_n | B_n, B_{1:n-1} ) || q_\phi ( b_n | B_n, B_{1:n-1} ) ]
\]
with respect to $\phi$, 
where $D_{KL}(\cdot||\cdot)$ is the Kullback-Leibler divergence and $p_\theta ( b_n | B_n, B_{1:n-1} ) = \frac{p_\theta ( B_n, b_n |    B_{1:n-1}  )}{ \int p_\theta ( B_n, b |    B_{1:n-1}  ) db  }$ is the intractable posterior distribution of $b_n$ from $p_\theta ( B_n, b_n |    B_{1:n-1}  )$.  To circumvent the intractability of the posterior distribution $p_\theta ( b_n | B_n, B_{1:n-1} )$, we again use the self-normalized importance sampling technique with the constructed neural network for $\log p_\theta(B_n,b_n|B_{1:n-1})$.  We estimate the negative gradient $-\nabla_{\phi} D_n$ in a similar way to the gradient estimation in \eqref{eq:generative_learn}:
\begin{equation}  \label{eq:inference_learn}
    - \nabla_{\phi} D_n \approx \sum_{j=1}^{ K_{\text{sp}} } \frac{ w_n^j }{  \sum_{j'=1}^{ K_{\text{sp}} } w_n^{j'}  } \nabla_{\phi} \log q_\phi (  b_n^j | B_n,   B_{1:n-1}  ),
\end{equation}
where the samples and importance sampling ratio $b_n^j$ and $w_n^j$  in \eqref{eq:generative_learn} can be used. (See Appendix \ref{sec:append_grad_estimate} for the full derivation.)

One issue regarding  the actual implementation of \eqref{eq:generative_learn} and \eqref{eq:inference_learn} is how to feed the current block information from $B_n$ into  $p_\theta$ and $q_\phi$. In the case of $q_\phi$ modeled as Gaussian distribution   $b_n \sim \mathcal{N}(\mu_{n}, \text{diag} ( \sigma_{n}^2 )  )$, the dependence on $B_n, B_{1:n-1}$ is through $\mu_n,\sigma_n$ which are functions of $h_n$. $h_n$ is a function of $Y_n^k$ and $h_{n-1}$, where $Y_n^k$ is a function $B_n$ and $h_{n-1}$ is a function of $B_{1:n-1}$, as seen in Fig. \ref{fig:proposed_architecture}. On the other hand, in the case of the neural network $\log p_\theta (B_n,b_n|B_{1:n-1})$, we choose to feed $Y_n^k$ and $b_n$. Note that $Y_n^k$ is a function of $B_n$,  $b_n$ is a function of $B_{1:n-1}$, and hence both are already conditioned on $B_{1:n-1}$.  

%%% RL learning
The part of RL learning is described as follows. 
With the sample $x_{t}=[a_{t-1};r_{t-1};o_{t}]$ and $\mu_n,\sigma_n$ from the learned $q_\phi$, the $z_t$-generation RNN is run to generate $z_t$ at each timestep $t$. The input $\mu_n,\sigma_n$ is common to the block, whereas $x_t$ changes at each timestep inside the block.  Then, the RL policy $\pi$ is learned using the sequence $\{(z_t,o_t,a_t,r_t),t=1,2,3,\cdots\}$ based on standard RL learning (either on-policy or off-policy), where $z_t$ is the side information compensating for the partiality in observation from the RL agent perspective.

The RL part and the model part are segregated by stopping the gradient from RL learning through $\mu_n$ and $\sigma_n$ to improve stability when training the RL agent \cite{han20vrm}. The pseudocode of the algorithm and the details are described in Appendix \ref{sec:append_algorithm} and \ref{sec:append_implementation}, respectively. Our source code is provided at \url{https://github.com/Giseung-Park/BlockSeq}.

% C: \ref{sec:append_algorithm}
% D: \ref{sec:append_implementation}

%%%%%%%%%%%%%%%%%%%%%%%%%%%%%%%%%%%%%%%%
\section{Experiments} \label{sec:experiments}
%%%%%%%%%%%%%%%%%%%%%%%%%%%%%%%%%%%%%%%%

In this section, we provide some numerical results to evaluate the proposed block model learning scheme for POMDPs.  In order to test the algorithm in various partially observable environments, we considered the following four types of partially observable environments:

\begin{itemize}
    \item Random noise is added to each state: Mountain Hike \cite{igl18dvrl}
    \item Some part of each state is missing: Pendulum - random missing version \cite{openai16gym,meng21memory}
    \item Memorizing long history is required: Sequential target-reaching task  \cite{han20sequential}
    \item Navigating agent cannot observe the whole map in maze: Minigrid \cite{gym_minigrid}
\end{itemize}

Note that the proposed method (denoted by Proposed) can be combined with any general RL algorithm. For the first three continuous action control tasks, we use the Soft Actor-Critic (SAC) algorithm \cite{haarnoja18sac}, which is an off-policy RL algorithm, as the background RL algorithm.  Then, we compare the performance of the proposed method with (i) SAC with raw observation input (SAC), (ii) SAC aided by the output of LSTM \cite{hochreiter97lstm}, a variant of RNN, driven by observation sequences (LSTM), (iii) VRM, which is a SAC-based method for partially observable continuous action control tasks \cite{han20vrm}, and (iv) RIMs as a drop-in replacement of LSTM. The $y$-axis in the three performance comparison figures represent the mean value of the returns of the most recent 100 episodes averaged over five random seeds. 

Since the Minigrid environment has discrete action spaces, we cannot use SAC and VRM, but instead, we use PPO \cite{schulman17ppo} as the background algorithm. Then the proposed algorithm is compared with PPO, PPO with LSTM, and PPO with RIMs over five seeds. (The details of the implementations are described in Appendix \ref{sec:append_implementation}.) 
% D: \ref{sec:append_implementation}

\subsection{Mountain Hike} \label{subsec:mthike}

% keepaspectratio,
\begin{figure}[!ht]
  \centering
    \subfigure[]{
    % \captionsetup{skip=50pt} % local setting for this subfigure
    % keepaspectratio,
    \includegraphics[width=.155\textwidth,height=0.17\textheight,valign=b]{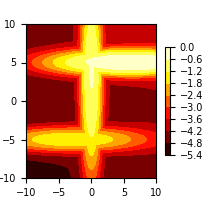}
    % \rule{0pt}{\normalbaselineskip} 
    % \bigskip
    % \vskip 0pt plus 1fil\relax
    % \vfill
    \label{fig:MTHike_env_map}
    }
    % \rule{0pt}{100pt} 
    \subfigure[]{ % width=.37\textwidth,height=0.147\textheight
    \includegraphics[width=.5\textwidth,height=0.17\textheight,keepaspectratio,valign=b]{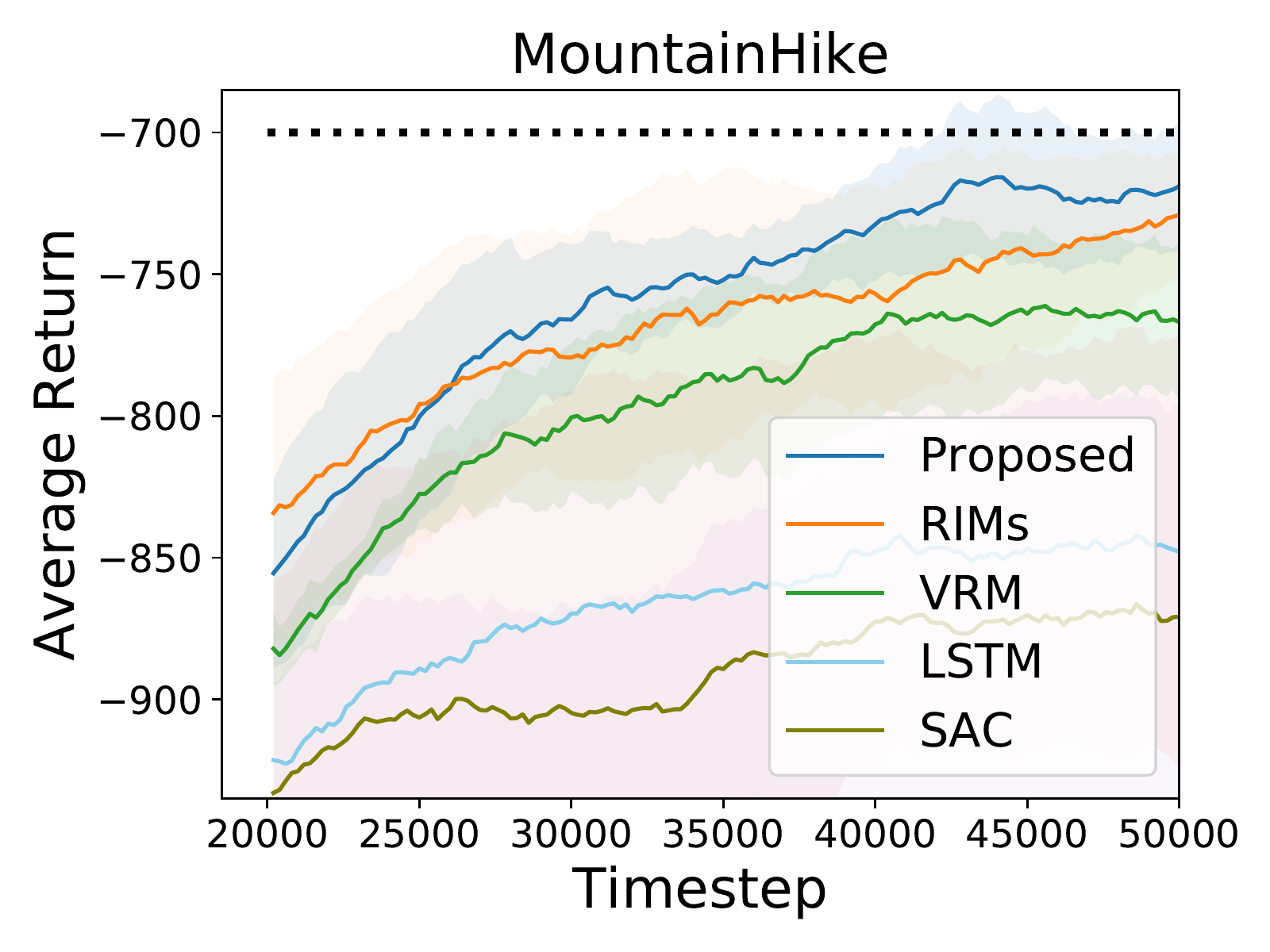}
    \label{fig:MTHike_performance}
    }
  \caption{(a) Distribution of reward function in the Mountain Hike environment   (b) Performance comparison in the Mountain Hike environment (The horizontal black dotted line shows the mean  performance of SAC over five seeds at 50000 timesteps when each observation is fully observable.)   }
  \label{fig:MTHike_results}
  \vspace{-2mm}
\end{figure}

The goal of the agent in Mountain Hike is to maximize the cumulative return by moving along the path of the high reward region, as shown in Fig. \ref{fig:MTHike_env_map}. Each state is a position of the agent, but the observation is received with the addition of Gaussian noise. (See Appendix \ref{sec:append_environment} for the details.) In Fig. \ref{fig:MTHike_performance}, it is seen that the proposed method outperforms the baselines in the Mountain Hike environment. The horizontal black dotted line shows the mean SAC performance over five seeds at 50000 steps without noise. Hence, the performance of the proposed method nearly approaches the SAC performance in the fully observable setting.
% E: \ref{sec:append_environment}

We applied Welch's t-test at the end of the training to statistically check the proposed method's gain over the baselines. This test is robust for comparison of different RL algorithms \cite{colas19welch}. Each $p$-value is the probability that the proposed algorithm does not outperform the compared baseline. Then the proposed algorithm outperforms the compared baseline with a $100(1-p) \%$ confidence level. The proposed method outperforms RIMs and VRM with 73  \% and 98 \% confidence levels, respectively.

\subsection{ Pendulum - Random Missing Version} \label{subsec:pendulum}

\begin{figure}[!ht]
   \centering
    \subfigure[]{
    \includegraphics[width=0.475\textwidth,height=0.15\textheight,keepaspectratio,valign=b]{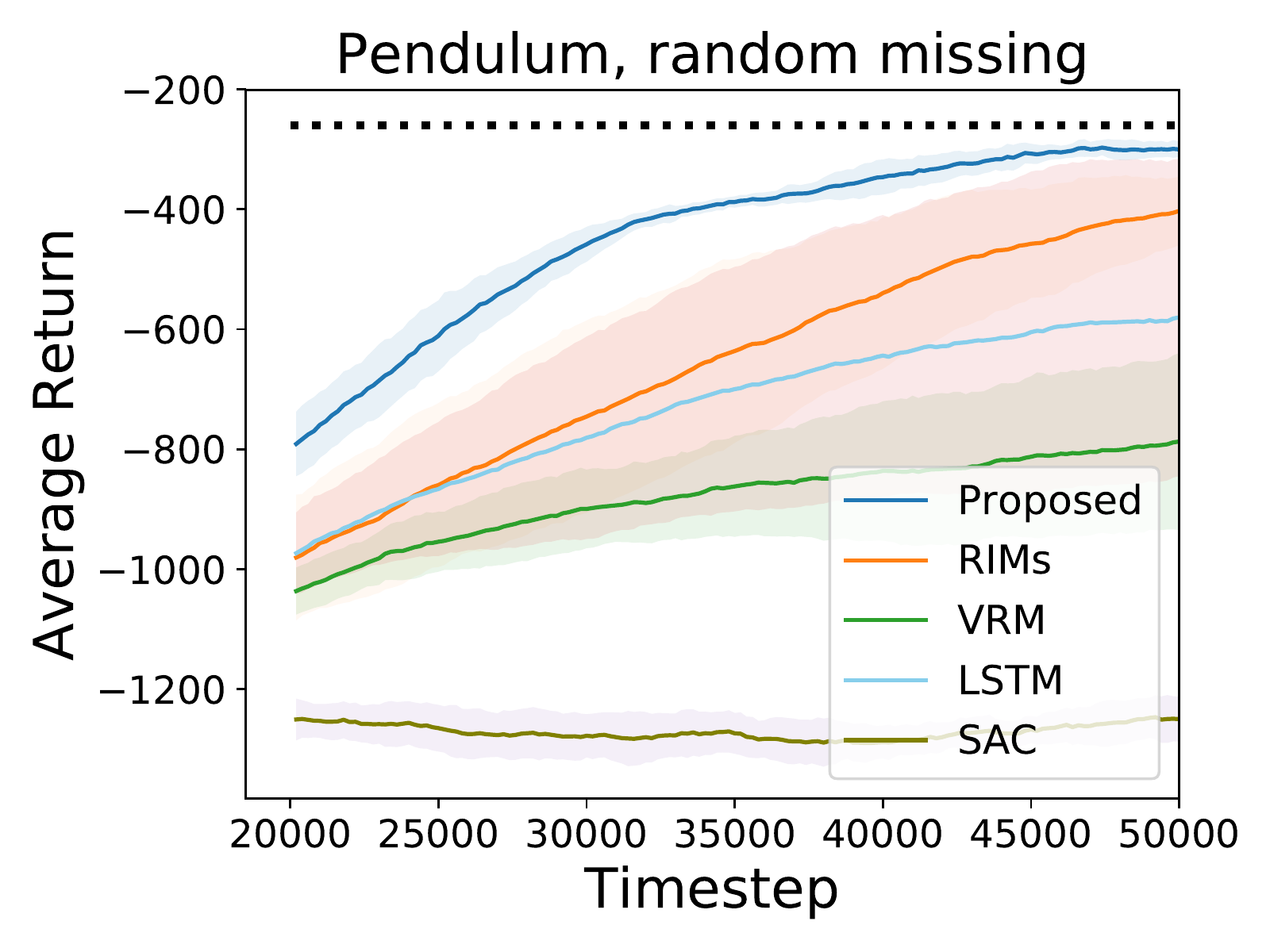}
    \label{fig:Pendulum_performance}
    }
    \subfigure[]{
    \begin{adjustbox}{width=.19\textwidth,height=0.06\textheight,valign=b} % 12/9. Change from 0.08 -> 0.06
        \begin{tabular}{ c  c   }
		\hline
		Method  & $p$-value \\ \hline
		Proposed  & - \\ 
		RIMs & $0.011$ \\ 
		VRM & $0.001$ \\ 
		LSTM &  $0.051$ \\
		SAC &  $4.57 \times 10^{-8}$ \\ 
		\hline
		\rule{0pt}{0.85\normalbaselineskip} 
	\end{tabular}
    \end{adjustbox}
    \label{fig:Pendulum_welch}
    }
  \caption{ (a) Performance comparison in the Pendulum environment (the horizontal black dotted line shows the mean  performance of SAC over five seeds at convergence when each observation is fully observable) and (b) $p$-values of the null hypothesis $H_0: \mu_{\text{proposed}}  \leq \mu_{\text{baseline}}$ based on  Welch's t-test, where $\mu_{\text{proposed}}$ and $ \mu_{\text{baseline}}$ are the performance means of the proposed method and each baseline, respectively}
  \label{fig:Pendulum_results}
%   \vspace{-2mm}
\end{figure}

We conducted experiments on the Pendulum control problem \cite{openai16gym}, where the pendulum is learned to swing up and stay upright during every episode. Unlike the original fully-observable version, each dimension of every state is converted to zero with probability $p_{\text{miss}}=0.1$ when the agent receives observation \cite{meng21memory}. This random missing setting induces partial observability and makes a simple control problem challenging. 

It is seen in Fig. \ref{fig:Pendulum_performance} that the proposed method outperforms the baselines.  The horizontal black dotted line shows the mean SAC performance at convergence when $p_{\text{miss}}=0.0$.  The performance of the proposed method nearly approaches the SAC performance in the fully observable setting, as seen in Fig. \ref{fig:Pendulum_performance}.  Fig. \ref{fig:Pendulum_welch} shows that the proposed method outperforms LSTM and RIMs with 95 \% and 99 \% confidence levels, respectively. 
Note in Fig. \ref{fig:Pendulum_performance} that the performance variance of the proposed method (which is 12.8) is significantly smaller than that of VRM and LSTM (147.6 and 264.4, respectively). This implies that the proposed method is learned more stably than the baselines.

\subsection{Sequential Target-reaching Task}
% keepaspectratio,

\begin{figure}[!ht]
  \centering
    \subfigure[]{
    \includegraphics[width=.165\textwidth,height=0.15\textheight,valign=b]{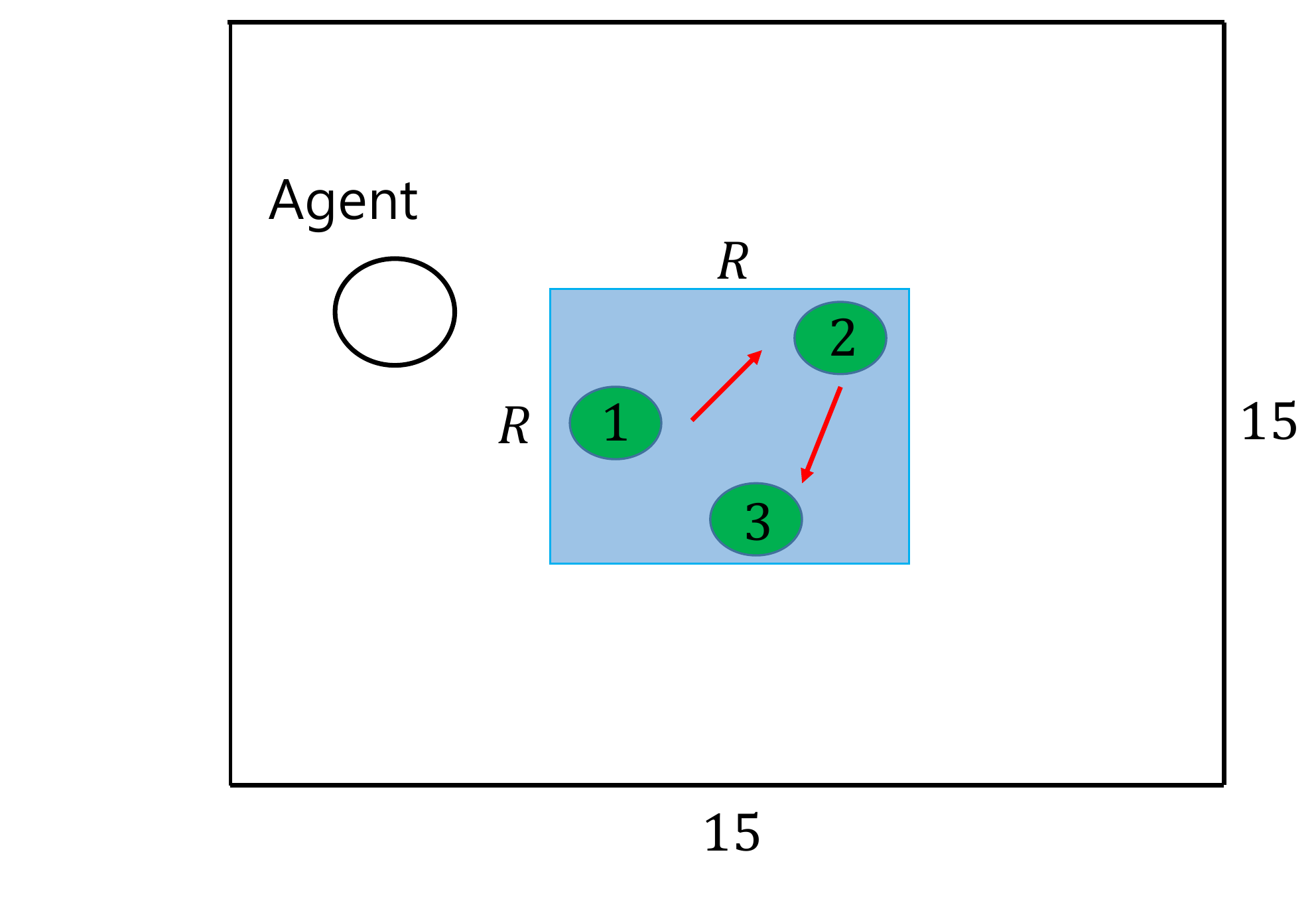}
    \label{fig:sequential_env_map}
    }
    \subfigure[]{
    \begin{adjustbox}{width=.28\textwidth,height=0.065\textheight,valign=b} % 12/9. Change from 0.08 -> 0.07; remove '|' between c's.
    \begin{tabular}{ c  c  c }
	\hline
	        & Success Rate & Success Rate \\ 
	Method  & $R=10$ ($\%$) & $R=15$ ($\%$) \\ 
	\hline
	Proposed & $\mathbf{98.4}$ & $\mathbf{91.4}$ \\ 
	RIMs & 32.4 & 4.2  \\ 
	VRM & 68.8 & 15.6  \\ 
	LSTM & 33.8 & 4.8    \\
	SAC & 1.2 & 0.0    \\ 
	\hline 
	\rule{0pt}{0.85\normalbaselineskip} % 12/9. added
    \end{tabular}
    \end{adjustbox}
    \label{tab:success_rate}
    }
    \subfigure[]{
    \includegraphics[width=.223\textwidth,keepaspectratio,valign=b]{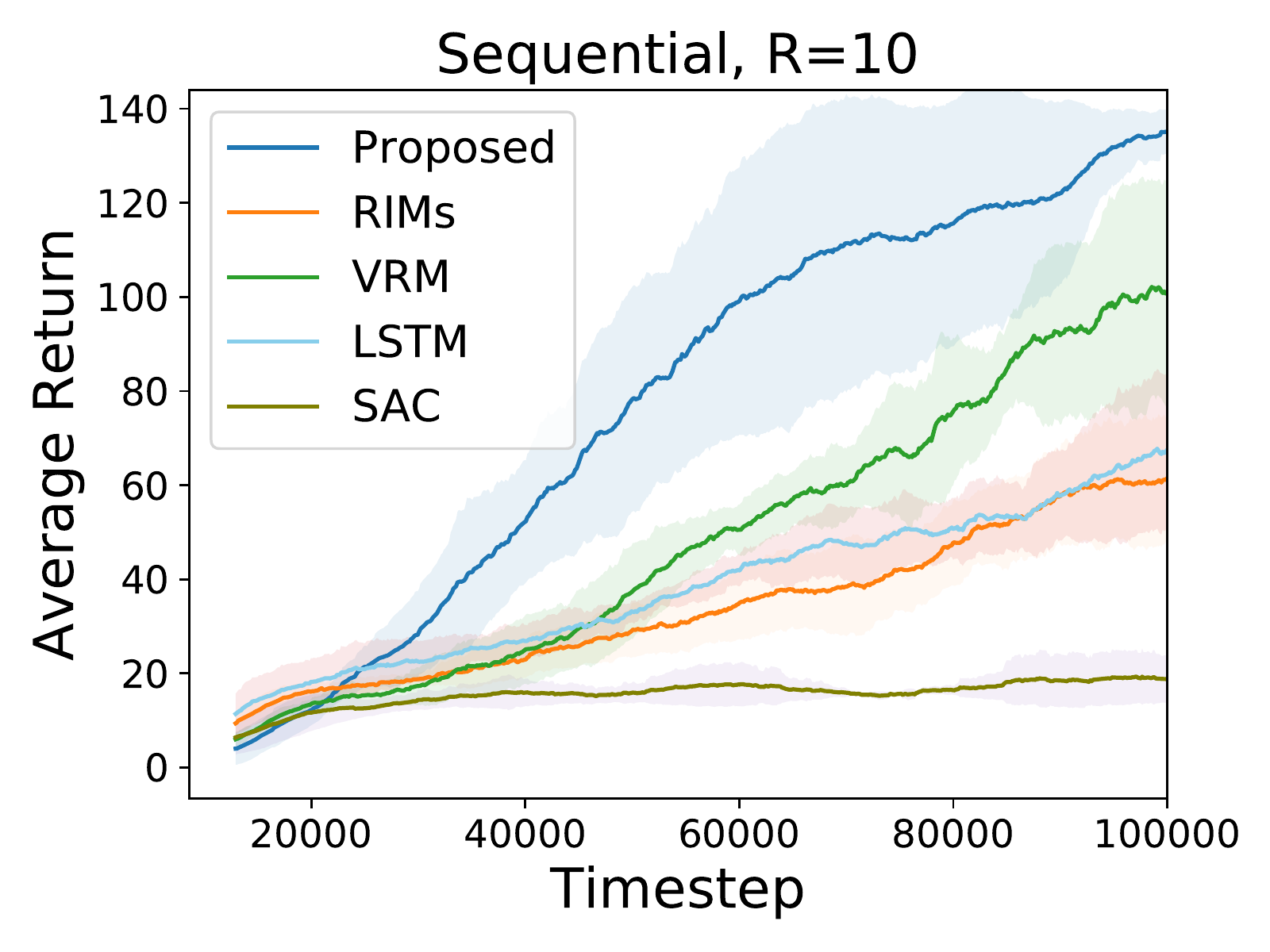}
    \label{fig:sequential_R10_results}
    }
    \subfigure[]{
    \includegraphics[width=.223\textwidth,keepaspectratio,valign=b]{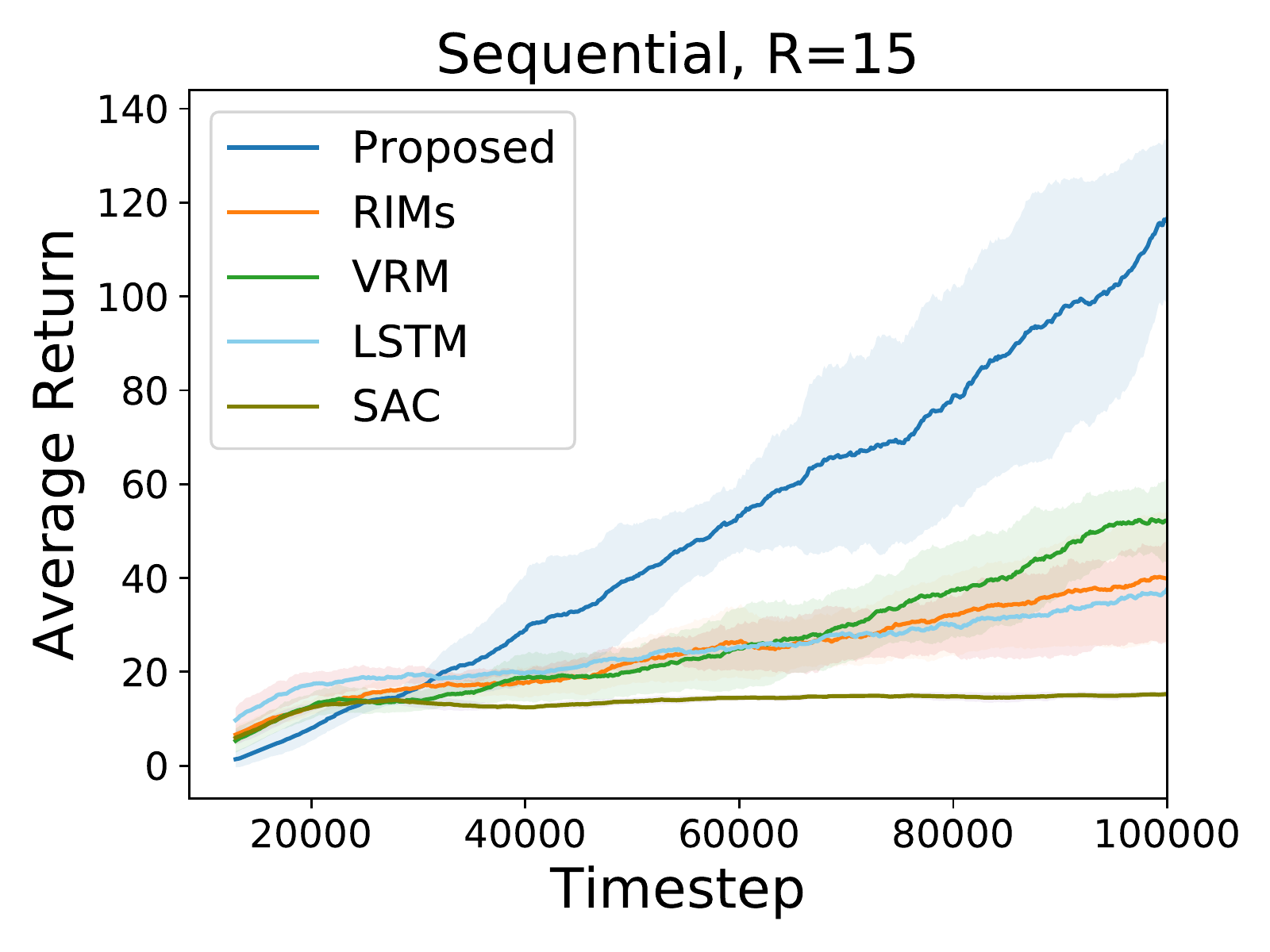}
    \label{fig:sequential_R15_results}
    }
  \caption{ (a) Sequential target-reaching task, (b) success rate in the task with $R=10$ and $R=15$, respectively, and (c) and (d) performance comparison with $R=10$ and $R=15$, respectively}
  \label{fig:sequential_results}
%   \vspace{-2mm}
\end{figure}

To verify that  the proposed model can learn long-term information effectively,  we conducted experiments in the sequential target-reaching task \cite{han20sequential}. The  sequential target-reaching task is shown in Fig. \ref{fig:sequential_env_map}. 
The agent has to visit three targets in order of $1 \rightarrow 2 \rightarrow 3$ as shown in Fig. \ref{fig:sequential_env_map}). Visiting the first target only yields $r_{\text{seq}}^1$, visiting the first and the second target yields $r_{\text{seq}}^2$, and visiting all the three target in order of  $1 \rightarrow 2 \rightarrow 3$ yields  $r_{\text{seq}}^3$, where $r_{\text{seq}}^3 > r_{\text{seq}}^2 > r_{\text{seq}}^1 > 0$.  Otherwise, the agent receives zero reward. When $R(0 < R \leq 15)$ increases, the distances among the three targets become larger, and the task becomes more challenging. The agent must memorize and properly use the past information to get the full reward.

In Figs. \ref{fig:sequential_R10_results} and \ref{fig:sequential_R15_results}, it is seen that the proposed method significantly outperforms the baselines.  Note that the performance gap between the proposed method and the baselines becomes large as the task becomes more difficult by increasing $R=10$ to $R=15$.
Welch's t-test shows that the proposed method outperforms VRM with 98 \% confidence level when $R=10$. The $p$-value compared to VRM is $2.84 \times 10^{-4}$ when $R=15$.

The success rate is an alternative measure other than the average return, removing the overestimation by reward function choice in the sequential target-reaching task. After training the block model and the RL agent, we loaded the trained models, evaluated 100 episodes for each model, and checked how many times the models successfully reached all three targets in the proper order. In Fig. \ref{tab:success_rate}, it is seen that the proposed method drastically outperforms the other baselines.

\subsection{Minigrid}

\begin{figure}[!ht]
  \centering
    %%% S9N3
    \subfigure[]{
    \includegraphics[width=.09\textwidth,height=0.13\textheight,valign=b]{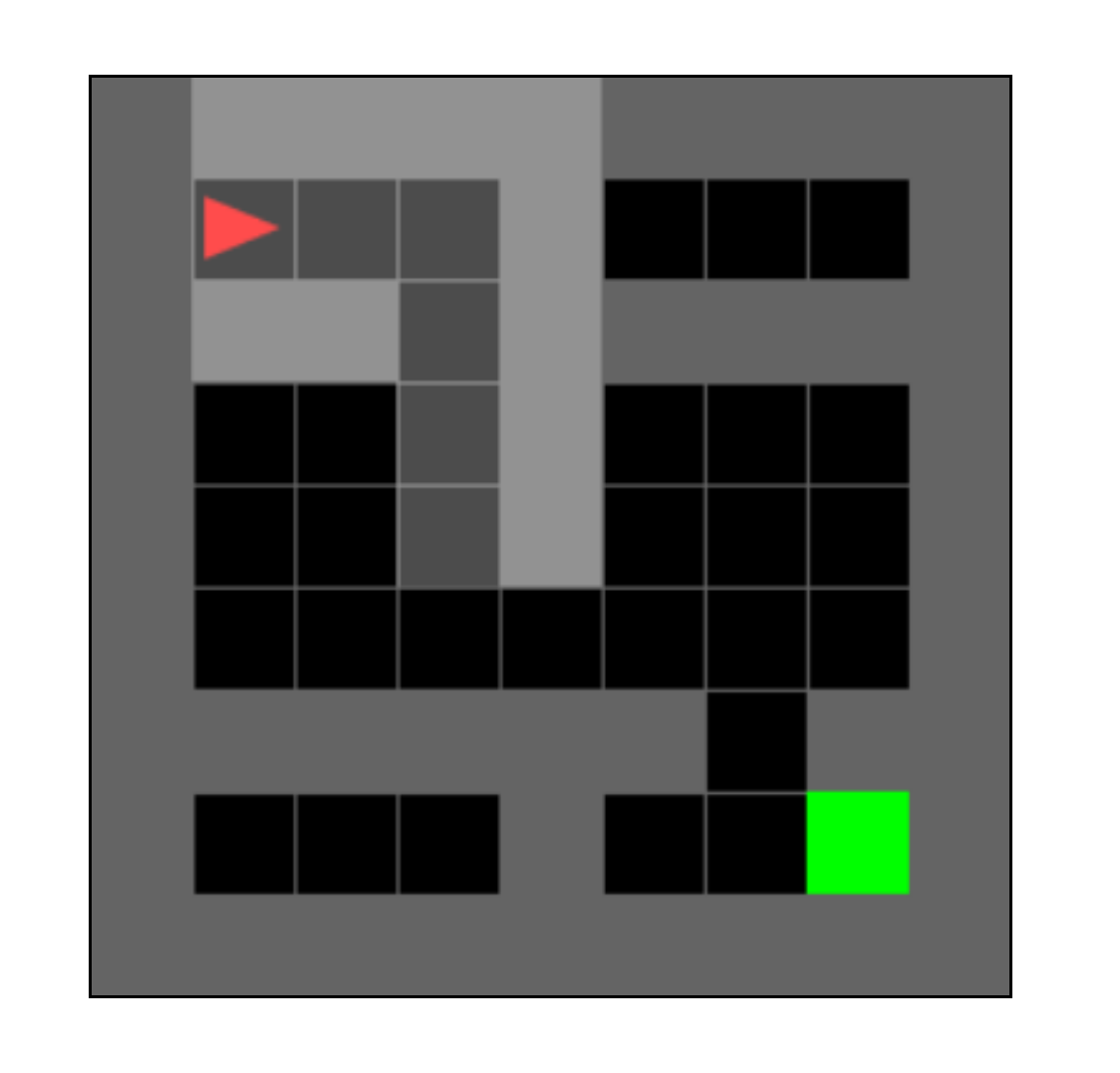}
    \label{fig:minigrid_S9N3_map}
    }
    \subfigure[]{
    \includegraphics[width=.32\textwidth,keepaspectratio,valign=b]{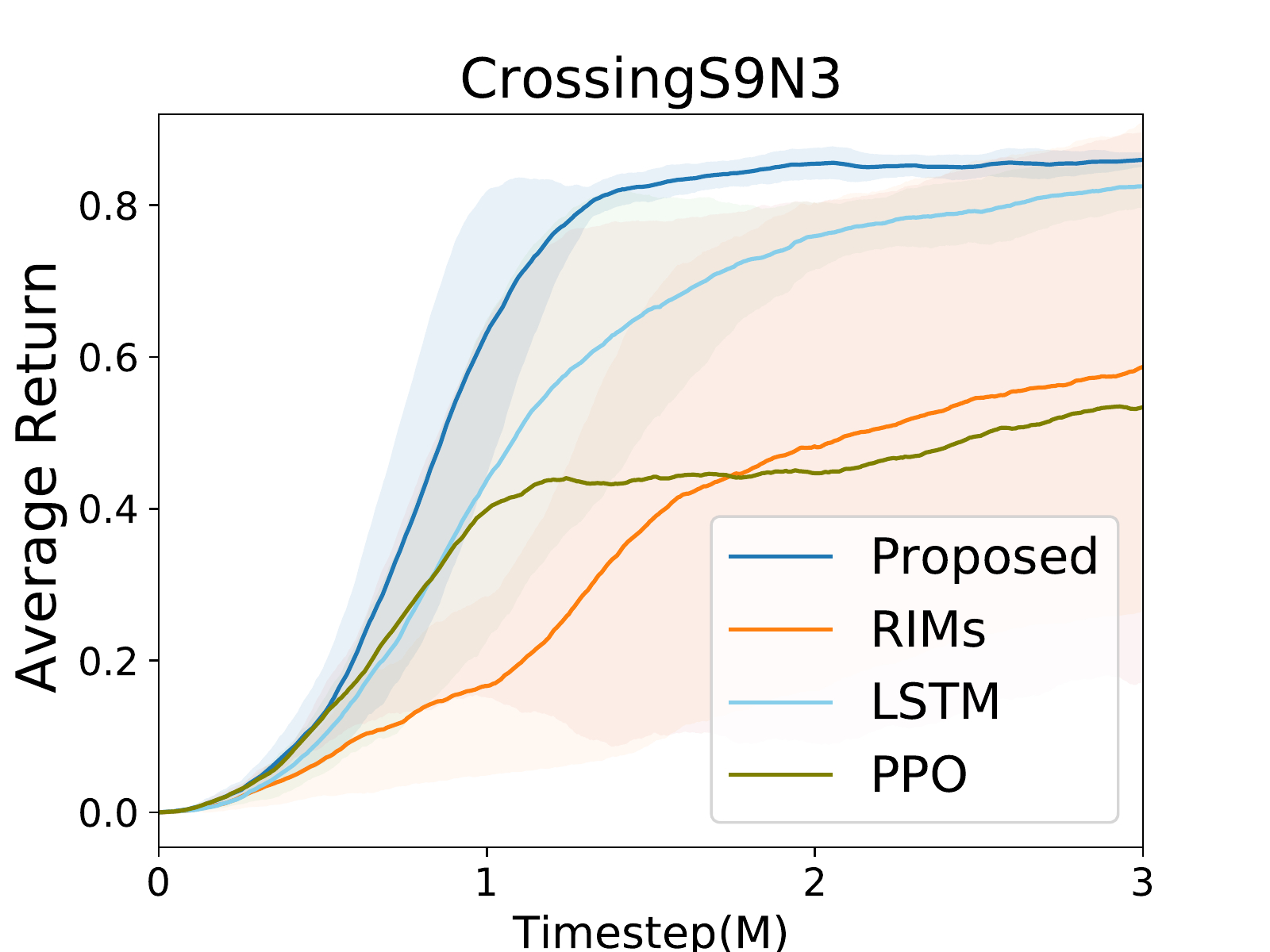}
    \label{fig:minigrid_S9N3_results}
    }
    \subfigure[]{
    \includegraphics[width=.1\textwidth,height=0.147\textheight,valign=b]{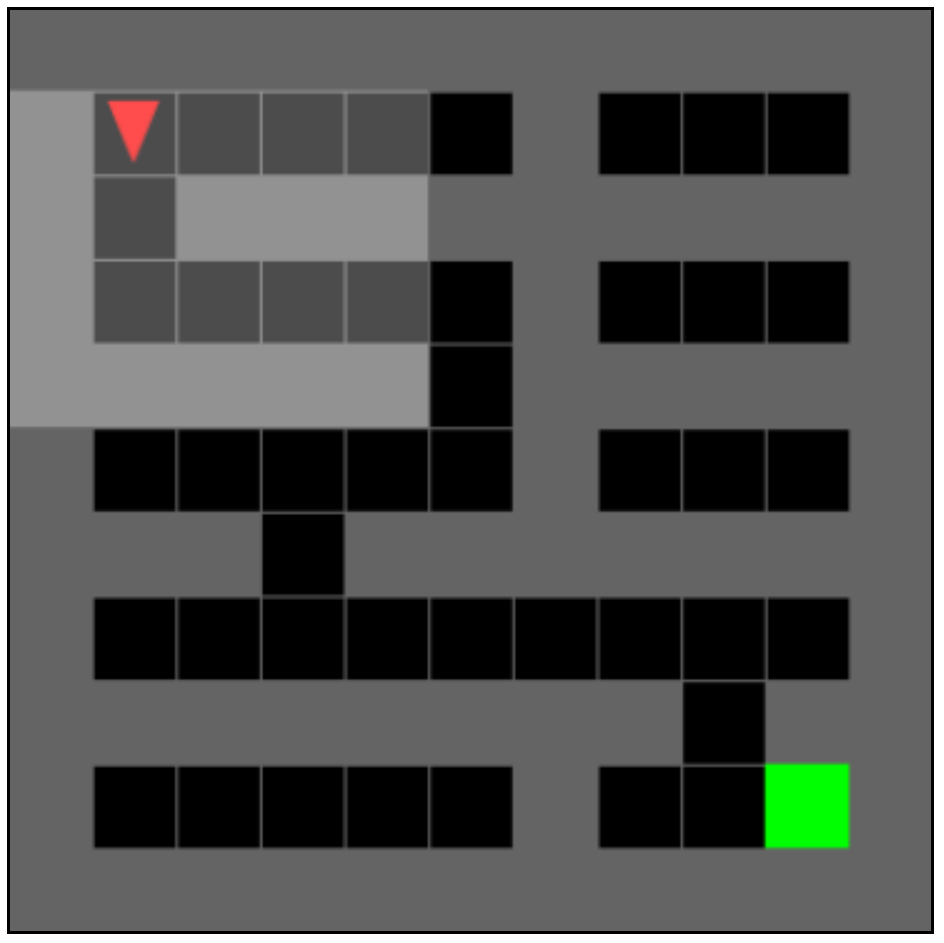}
    \label{fig:minigrid_S11N5_map}
    }
    \subfigure[]{
    \includegraphics[width=.32\textwidth,keepaspectratio,valign=b]{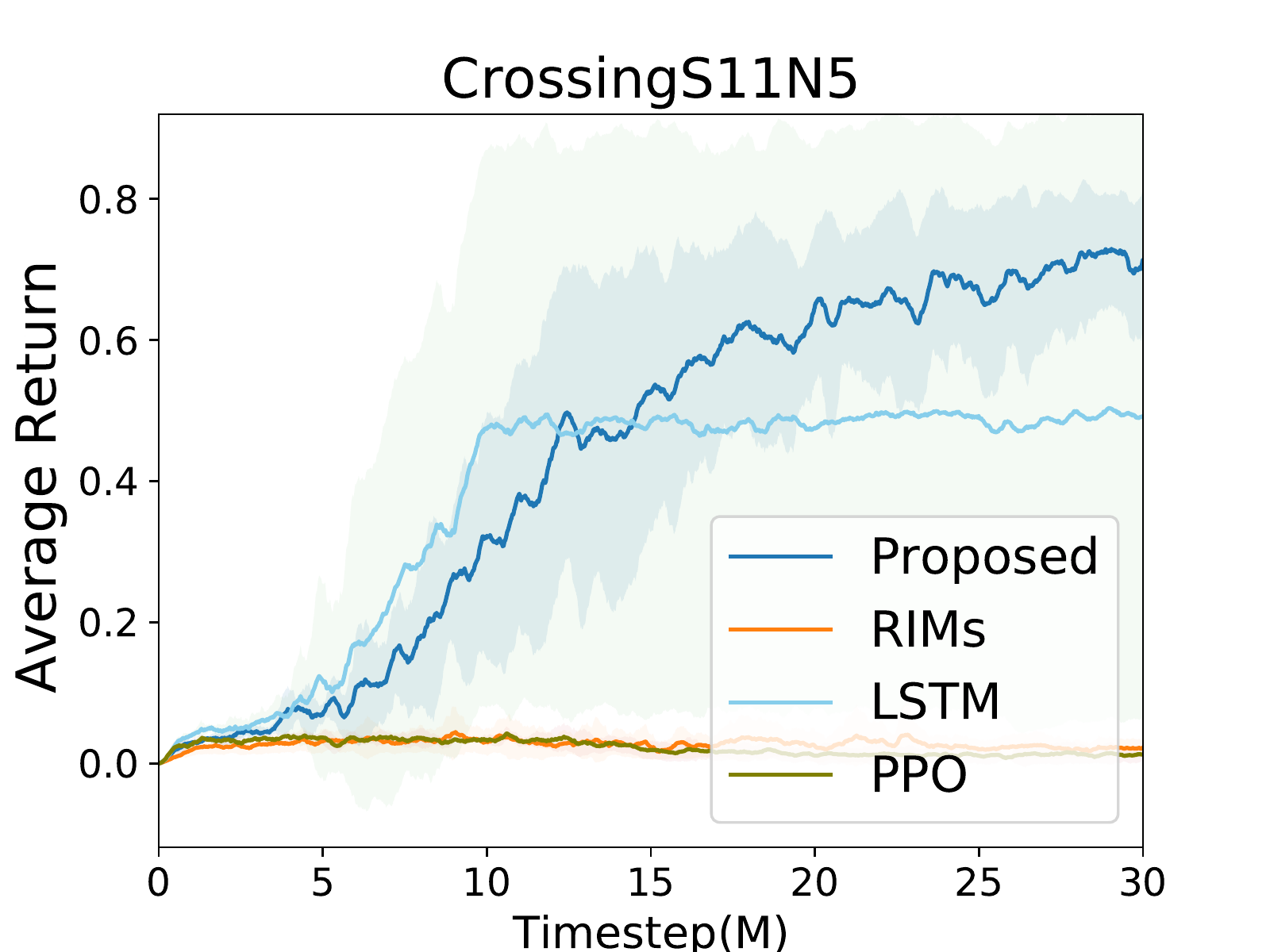}
    \label{fig:minigrid_S11N5_results}
    }
  \caption{  (a) A sample map of Minigrid CrossingS9N3 task and (b) performance comparison in CrossingS9N3. (c) A sample map of Minigrid CrossingS11N5 task and (d) performance comparison in CrossingS11N5. The $y$-axis represents the mean value of the returns of the most recent two hundred iterations over the five seeds   }
  \label{fig:minigrid_results}
  \vspace{-4mm}
\end{figure}

We considered partially observable maze navigation environments with sparse reward, as seen in Figs. \ref{fig:minigrid_S9N3_map} and \ref{fig:minigrid_S11N5_map} \cite{gym_minigrid}. The agent (red triangle) receives a nonzero reward $1 - 0.9 \frac{t_{\text{step}}}{N_{\text{max}}}$ only when it reaches the green square, where $N_{\text{max}}$ is the maximum episode length and $t_{\text{step}} (\leq  N_{\text{max}}) $ is the total timestep before success. Otherwise, the agent receives zero reward.  A new map with the same size but different shapes is generated at every episode, and the agent starts to navigate again. The agent must learn to cross the narrow path with partial observation and sparse reward. (See Appendix \ref{sec:append_environment} for more details.) 
% E: \ref{sec:append_environment}

In Figs. \ref{fig:minigrid_S9N3_results} and \ref{fig:minigrid_S11N5_results}, it is seen that the proposed method outperforms the considered baselines even when the size of map increases and the difficulty becomes higher. According to Welch’s t-test, the proposed method outperforms PPO with RIMs (RIMs), PPO with LSTM (LSTM), and PPO in \ref{fig:minigrid_S9N3_results} with $93 \%, 98 \%$, and $94 \%$, respectively. The  $p$-value of LSTM in Fig. \ref{fig:minigrid_S11N5_results} is 0.159.

%%%%%%%%%

\section{ Ablation Study  } \label{sec:ablation_study}

Recall that the proposed block model consists of the blockwise RNN and self-attention. In Section \ref{subsec:ablation_components_compare}, we investigate the contribution to the performance improvement of the blockwise RNN and the self-attention.  In Section \ref{subsec:ablation_compression_compare}, we replace the proposed compression method with other methods while using the same self-attention. (See Appendix \ref{sec:append_ablation_hyperparameter} for the effect of hyperparameters $L$ and $k$.)
% F: \ref{sec:append_ablation_hyperparameter}

\subsection{Effect of Components}  \label{subsec:ablation_components_compare}

We include the method using only self-attention without blockwise RNN (denoted by `Self-attention only'). $Y_n^k \in \mathbb{R}^{k \cdot d}$, a single vector from concatenation of $k$ selected elements, is fed into RL agent instead of $\mu_n$ and $\sigma_n$. The self-attention is trained end-to-end with the RL agent.

We also add the method using only blockwise RNN without self-attention (`Blockwise RNN only') by replacing the self-attention with a feedforward neural network (FNN). The replaced FNN maps each $d$-dimensional input $x_t \in B_n$ in a block to an $S_{\text{FNN}}$-dimensional vector. Instead of $Y_n^k$, $(L \cdot S_{\text{FNN}})$-dimensional transformed block is used for the blockwise RNN input. For fair comparison, we set $S_{\text{FNN}}$ such that $L \cdot S_{\text{FNN}}$ is equal to the dimension of $Y_n^k(=k \cdot d)$. 

In Tab. \ref{tab:ablation_study_rebuttal}, we observe that blockwise RNN plays an essential role for performance improvement in both sequential target-reaching task and Pendulum. In Pendulum, the effect of self-attention is critical for performance improvement. 

\begin{center}
\begin{table}[!ht]
\centering
\begin{tabular}{ c | c | c  }
	\hline
	                 & Success Rate  & Average Return  \\ 
	      Method     & $R=15$ ($\%$)  & in Pendulum  \\ 
    %   Method     & $R=15$ ($\%$)  &   \\ 
	\hline
	Proposed  & $\mathbf{91.4}$ & -$\mathbf{300.2}$ \\ 
	Self-attention only & 21.4 & -342.9  \\ 
	Blockwise RNN only & 90.8 &  -467.7 \\ 
	Best baseline & 15.6 (VRM) & -402.7 (RIMs) \\ 
	\hline 
\end{tabular}
\caption{ \label{tab:ablation_study_rebuttal}  Ablation on effect of components: Sequential target-reaching task with $R=15$ (middle) and Pendulum (right) }
\end{table}
\vspace{-0.6cm}
\end{center}

%%%%%%%%%%%%%%%%%%%%%%%%%%%%

\subsection{ Effect of Compression Methods } \label{subsec:ablation_compression_compare}

\begin{center}
\begin{table}[!ht]
\centering
\begin{tabular}{ c | c  c  }
	\hline
	        Method   & Average Return  & $p$-value  \\ 
	\hline
	Proposed  & -$\mathbf{300.2}$ & - \\ 
	Pooling & -354.1 & 0.001  \\ 
	Top-$k$ Average & -352.0 &  0.011 \\ 
	Linear & -346.5 & 0.007 \\ 
	Random & -349.6 & 0.024 \\ 
	\hline 
\end{tabular}
\caption{ \label{tab:ablation_study_compress}  Performance comparison with other compression methods in the considered Pendulum environment }
\end{table}
\vspace{-0.6cm}
\end{center}

To check the effectiveness of the proposed compression method which selects $Y_n^k$ from $Y_n$, we conducted performance comparison with other compression methods in the Pendulum environment. Instead of using $Y_n^k$ as an input to the blockwise RNN in the proposed method, the considered baselines use (i) $\sum_{i=(n-1)L + 1}^{nL} y_i$ (Pooling), (ii) averaging over the $k$ selected elements in $Y_n$ weighted by normalized corresponding contributions (Top-$k$ Average), (iii) $[ M_{\text{comp}} y_{(n-1)L + 1}; \cdots ;  M_{\text{comp}} y_{nL}  ]$ with a trainable matrix $M_{\text{comp}} \in \mathbb{R}^{\frac{kd}{L} \times d}$ (Linear), or (iv) $k$ randomly chosen elements in $Y_n$ (Random). 

The comparison result is shown in Tab. \ref{tab:ablation_study_compress}. In Pendulum, self-attention has effective compression ability since all the considered compression methods using self-attention outperform the `Blockwise RNN only' method (with average -467.7) in Section  \ref{subsec:ablation_components_compare}. Among the considered compression methods with self-attention, the proposed method induces the least relevant information loss.

\section{Conclusion}

In this paper, we have proposed a new blockwise sequential model learning for POMDPs. The proposed model compresses the input sample sequences using self-attention for each data block and passes the compressed information to the next block using RNN. The compressed information from the block model is fed into the RL agent with the corresponding data block to improve the RL performance in POMDPs. The proposed architecture is learned based on direct gradient estimation using self-normalized importance sampling, making the learning efficient. By exploiting the advantages of self-attention and RNN, the proposed method outperforms the previous approaches to POMDPs in the considered partially observable environments.

\newpage

\section*{Acknowledgments}
This research was supported by Basic Science Research Program through the National Research Foundation of Korea (NRF) funded by the Ministry of Science, ICT \& Future Planning (NRF-2021R1A2C2009143).

\bibliography{aaai22}

\begin{thebibliography}{26}
\providecommand{\natexlab}[1]{#1}

\bibitem[{Arjovsky, Chintala, and Bottou(2017)}]{arjovsky17wgan}
Arjovsky, M.; Chintala, S.; and Bottou, L. 2017.
\newblock Wasserstein Generative Adversarial Networks.
\newblock In Precup, D.; and Teh, Y.~W., eds., \emph{Proceedings of the 34th
  International Conference on Machine Learning, {ICML} 2017, Sydney, NSW,
  Australia, 6-11 August 2017}, volume~70 of \emph{Proceedings of Machine
  Learning Research}, 214--223. {PMLR}.

\bibitem[{Ba, Kiros, and Hinton(2016)}]{ba16layernorm}
Ba, L.~J.; Kiros, J.~R.; and Hinton, G.~E. 2016.
\newblock Layer Normalization.
\newblock \emph{CoRR}, abs/1607.06450.

\bibitem[{Bornschein and Bengio(2015)}]{bornschein14rws}
Bornschein, J.; and Bengio, Y. 2015.
\newblock Reweighted Wake-Sleep.
\newblock In Bengio, Y.; and LeCun, Y., eds., \emph{3rd International
  Conference on Learning Representations, {ICLR} 2015, San Diego, CA, USA, May
  7-9, 2015, Conference Track Proceedings}.

\bibitem[{Brockman et~al.(2016)Brockman, Cheung, Pettersson, Schneider,
  Schulman, Tang, and Zaremba}]{openai16gym}
Brockman, G.; Cheung, V.; Pettersson, L.; Schneider, J.; Schulman, J.; Tang,
  J.; and Zaremba, W. 2016.
\newblock OpenAI Gym.

\bibitem[{Chevalier-Boisvert, Willems, and Pal(2018)}]{gym_minigrid}
Chevalier-Boisvert, M.; Willems, L.; and Pal, S. 2018.
\newblock Minimalistic Gridworld Environment for OpenAI Gym.
\newblock \url{https://github.com/maximecb/gym-minigrid}.

\bibitem[{Cho et~al.(2014)Cho, van Merrienboer, G{\"{u}}l{\c{c}}ehre, Bahdanau,
  Bougares, Schwenk, and Bengio}]{cho14gru}
Cho, K.; van Merrienboer, B.; G{\"{u}}l{\c{c}}ehre, {\c{C}}.; Bahdanau, D.;
  Bougares, F.; Schwenk, H.; and Bengio, Y. 2014.
\newblock Learning Phrase Representations using {RNN} Encoder-Decoder for
  Statistical Machine Translation.
\newblock In Moschitti, A.; Pang, B.; and Daelemans, W., eds.,
  \emph{Proceedings of the 2014 Conference on Empirical Methods in Natural
  Language Processing, {EMNLP} 2014, October 25-29, 2014, Doha, Qatar, {A}
  meeting of SIGDAT, a Special Interest Group of the {ACL}}, 1724--1734. {ACL}.

\bibitem[{Chung et~al.(2015)Chung, Kastner, Dinh, Goel, Courville, and
  Bengio}]{chung15vrnn}
Chung, J.; Kastner, K.; Dinh, L.; Goel, K.; Courville, A.~C.; and Bengio, Y.
  2015.
\newblock A Recurrent Latent Variable Model for Sequential Data.
\newblock In Cortes, C.; Lawrence, N.~D.; Lee, D.~D.; Sugiyama, M.; and
  Garnett, R., eds., \emph{Advances in Neural Information Processing Systems
  28: Annual Conference on Neural Information Processing Systems 2015, December
  7-12, 2015, Montreal, Quebec, Canada}, 2980--2988.

\bibitem[{Colas, Sigaud, and Oudeyer(2019)}]{colas19welch}
Colas, C.; Sigaud, O.; and Oudeyer, P. 2019.
\newblock A Hitchhiker's Guide to Statistical Comparisons of Reinforcement
  Learning Algorithms.
\newblock In \emph{Reproducibility in Machine Learning, {ICLR} 2019 Workshop,
  New Orleans, Louisiana, United States, May 6, 2019}. OpenReview.net.

\bibitem[{Goyal et~al.(2021)Goyal, Lamb, Hoffmann, Sodhani, Levine, Bengio, and
  Sch{\"{o}}lkopf}]{goyal21rims}
Goyal, A.; Lamb, A.; Hoffmann, J.; Sodhani, S.; Levine, S.; Bengio, Y.; and
  Sch{\"{o}}lkopf, B. 2021.
\newblock Recurrent Independent Mechanisms.
\newblock In \emph{9th International Conference on Learning Representations,
  {ICLR} 2021, Virtual Event, Austria, May 3-7, 2021}. OpenReview.net.

\bibitem[{Haarnoja et~al.(2018)Haarnoja, Zhou, Abbeel, and
  Levine}]{haarnoja18sac}
Haarnoja, T.; Zhou, A.; Abbeel, P.; and Levine, S. 2018.
\newblock Soft Actor-Critic: Off-Policy Maximum Entropy Deep Reinforcement
  Learning with a Stochastic Actor.
\newblock In Dy, J.~G.; and Krause, A., eds., \emph{Proceedings of the 35th
  International Conference on Machine Learning, {ICML} 2018,
  Stockholmsm{\"{a}}ssan, Stockholm, Sweden, July 10-15, 2018}, volume~80 of
  \emph{Proceedings of Machine Learning Research}, 1856--1865. {PMLR}.

\bibitem[{Han, Doya, and Tani(2020{\natexlab{a}})}]{han20sequential}
Han, D.; Doya, K.; and Tani, J. 2020{\natexlab{a}}.
\newblock Self-organization of action hierarchy and compositionality by
  reinforcement learning with recurrent neural networks.
\newblock \emph{Neural Networks}, 129: 149--162.

\bibitem[{Han, Doya, and Tani(2020{\natexlab{b}})}]{han20vrm}
Han, D.; Doya, K.; and Tani, J. 2020{\natexlab{b}}.
\newblock Variational Recurrent Models for Solving Partially Observable Control
  Tasks.
\newblock In \emph{8th International Conference on Learning Representations,
  {ICLR} 2020, Addis Ababa, Ethiopia, April 26-30, 2020}. OpenReview.net.

\bibitem[{Han et~al.(2021)Han, Min, Han, Li, and Zhang}]{han2021disentangled}
Han, J.; Min, M.~R.; Han, L.; Li, L.~E.; and Zhang, X. 2021.
\newblock Disentangled Recurrent Wasserstein Autoencoder.
\newblock In \emph{International Conference on Learning Representations}.

\bibitem[{Hausknecht and Stone(2015)}]{hausknecht15drqn}
Hausknecht, M.~J.; and Stone, P. 2015.
\newblock Deep Recurrent Q-Learning for Partially Observable MDPs.
\newblock In \emph{2015 {AAAI} Fall Symposia, Arlington, Virginia, USA,
  November 12-14, 2015}, 29--37. {AAAI} Press.

\bibitem[{Hochreiter and Schmidhuber(1997)}]{hochreiter97lstm}
Hochreiter, S.; and Schmidhuber, J. 1997.
\newblock Long Short-Term Memory.
\newblock \emph{Neural Comput.}, 9(8): 1735--1780.

\bibitem[{Igl et~al.(2018)Igl, Zintgraf, Le, Wood, and Whiteson}]{igl18dvrl}
Igl, M.; Zintgraf, L.; Le, T.~A.; Wood, F.; and Whiteson, S. 2018.
\newblock Deep Variational Reinforcement Learning for {POMDP}s.
\newblock In Dy, J.; and Krause, A., eds., \emph{Proceedings of the 35th
  International Conference on Machine Learning}, volume~80 of \emph{Proceedings
  of Machine Learning Research}, 2117--2126. PMLR.

\bibitem[{Le et~al.(2018)Le, Igl, Rainforth, Jin, and Wood}]{le18aesmc}
Le, T.~A.; Igl, M.; Rainforth, T.; Jin, T.; and Wood, F. 2018.
\newblock Auto-Encoding Sequential Monte Carlo.
\newblock In \emph{6th International Conference on Learning Representations,
  {ICLR} 2018, Vancouver, BC, Canada, April 30 - May 3, 2018, Conference Track
  Proceedings}. OpenReview.net.

\bibitem[{Le et~al.(2019)Le, Kosiorek, Siddharth, Teh, and
  Wood}]{le19revisitrws}
Le, T.~A.; Kosiorek, A.~R.; Siddharth, N.; Teh, Y.~W.; and Wood, F. 2019.
\newblock Revisiting Reweighted Wake-Sleep for Models with Stochastic Control
  Flow.
\newblock In Globerson, A.; and Silva, R., eds., \emph{Proceedings of the
  Thirty-Fifth Conference on Uncertainty in Artificial Intelligence, {UAI}
  2019, Tel Aviv, Israel, July 22-25, 2019}, volume 115 of \emph{Proceedings of
  Machine Learning Research}, 1039--1049. {AUAI} Press.

\bibitem[{Li and Mandt(2018)}]{li18seqae}
Li, Y.; and Mandt, S. 2018.
\newblock Disentangled Sequential Autoencoder.
\newblock In Dy, J.~G.; and Krause, A., eds., \emph{Proceedings of the 35th
  International Conference on Machine Learning, {ICML} 2018,
  Stockholmsm{\"{a}}ssan, Stockholm, Sweden, July 10-15, 2018}, volume~80 of
  \emph{Proceedings of Machine Learning Research}, 5656--5665. {PMLR}.

\bibitem[{Maddison et~al.(2017)Maddison, Lawson, Tucker, Heess, Norouzi, Mnih,
  Doucet, and Teh}]{maddison17fivo}
Maddison, C.~J.; Lawson, D.; Tucker, G.; Heess, N.; Norouzi, M.; Mnih, A.;
  Doucet, A.; and Teh, Y.~W. 2017.
\newblock Filtering Variational Objectives.
\newblock In Guyon, I.; von Luxburg, U.; Bengio, S.; Wallach, H.~M.; Fergus,
  R.; Vishwanathan, S. V.~N.; and Garnett, R., eds., \emph{Advances in Neural
  Information Processing Systems 30: Annual Conference on Neural Information
  Processing Systems 2017, December 4-9, 2017, Long Beach, CA, {USA}},
  6573--6583.

\bibitem[{Meng, Gorbet, and Kulic(2021)}]{meng21memory}
Meng, L.; Gorbet, R.; and Kulic, D. 2021.
\newblock Memory-based Deep Reinforcement Learning for {POMDP}.
\newblock \emph{CoRR}, abs/2102.12344.

\bibitem[{Naesseth et~al.(2018)Naesseth, Linderman, Ranganath, and
  Blei}]{naesseth18vsmc}
Naesseth, C.~A.; Linderman, S.~W.; Ranganath, R.; and Blei, D.~M. 2018.
\newblock Variational Sequential Monte Carlo.
\newblock In Storkey, A.~J.; and P{\'{e}}rez{-}Cruz, F., eds.,
  \emph{International Conference on Artificial Intelligence and Statistics,
  {AISTATS} 2018, 9-11 April 2018, Playa Blanca, Lanzarote, Canary Islands,
  Spain}, volume~84 of \emph{Proceedings of Machine Learning Research},
  968--977. {PMLR}.

\bibitem[{Richter and Wattenhofer(2020)}]{richter20normattention}
Richter, O.; and Wattenhofer, R. 2020.
\newblock Normalized Attention Without Probability Cage.
\newblock \emph{CoRR}, abs/2005.09561.

\bibitem[{Schulman et~al.(2017)Schulman, Wolski, Dhariwal, Radford, and
  Klimov}]{schulman17ppo}
Schulman, J.; Wolski, F.; Dhariwal, P.; Radford, A.; and Klimov, O. 2017.
\newblock Proximal Policy Optimization Algorithms.
\newblock \emph{CoRR}, abs/1707.06347.

\bibitem[{Vaswani et~al.(2017)Vaswani, Shazeer, Parmar, Uszkoreit, Jones,
  Gomez, Kaiser, and Polosukhin}]{vaswani17attention}
Vaswani, A.; Shazeer, N.; Parmar, N.; Uszkoreit, J.; Jones, L.; Gomez, A.~N.;
  Kaiser, L.; and Polosukhin, I. 2017.
\newblock Attention is All you Need.
\newblock In Guyon, I.; von Luxburg, U.; Bengio, S.; Wallach, H.~M.; Fergus,
  R.; Vishwanathan, S. V.~N.; and Garnett, R., eds., \emph{Advances in Neural
  Information Processing Systems 30: Annual Conference on Neural Information
  Processing Systems 2017, December 4-9, 2017, Long Beach, CA, {USA}},
  5998--6008.

\bibitem[{Zhu, Li, and Poupart(2017)}]{zhu17adrqn}
Zhu, P.; Li, X.; and Poupart, P. 2017.
\newblock On Improving Deep Reinforcement Learning for POMDPs.
\newblock \emph{CoRR}, abs/1704.07978.

\end{thebibliography}

%%%%%%%%%%%%%%%%%%%%% Appendix %%%%%%%%%%%%%%%%%%%%%%%%%%

\newpage

\appendix
\onecolumn

\newpage
\section*{ Notations }

\begin{table}[!ht]
	\centering
	\begin{adjustbox}{width=\textwidth}
		\begin{tabular}{ll}
			% \begin{tabular}{p{2.2cm} p{0.7cm} p{0.7cm} p{0.7cm} p{0.7cm}}
			\toprule
			%\multicolumn{7}{c}{}                   \\
			%\cmidrule{2-7}
			Notations & Descriptions   \\
			%\midrule
			\hline
			$p_\theta$ & Generative model \\
			$q_\phi$ & Inference model  \\
			$x_t$ & Input data $x_{t} = [a_{t-1}; r_{t-1}; o_{t}  ]$ the column-wise concatenation of \\ & action $a$, reward $r$, and the next (partial) observation $o$  \\
			$d$  & Dimension of each input data $x_t$ \\
			$T$ & Length of sample sequence $\{ x_t \}_{t=1}^T$  \\
			$L$ & Block length $(>1)$ \\
			$N$ & Number of blocks given $\{ x_t \}_{t=1}^T$ satisfying $T = NL$. \\
			$B_n$  & $n$-th block: $x_{(n-1)L+1:nL} = [x_{(n-1)L+1}, x_{(n-1)L+2}, \cdots, x_{nL}]^T$. $B = B_1$  \\
% 			$B$  & $B = B_1$   \\
			$b_n$ & $n$-th (stochastic) block latent variable \\
			$Z_n$ & Input variables $z_{(n-1)L+1 : nL}$ to the RL agent at the $n$-th block \\
			$h_n$ & Latent variable of RNN in the block model containing information of  $B_{1:n}$ \\
			$\mu_n, \sigma_n$ & Mean and diagonal standard deviation vectors of $q_{\phi} (\cdot | B_{1:n}) = \mathcal{N}(\mu_{n}, \text{diag} ( \sigma_{n}^2 )  )$ \\
			$Y_n$ & Output sequence $y_{(n-1)L+1:nL}$ of self-attention given $B_n$ as input. $Y = Y_1$   \\
% 			$Y$ & $Y = Y_1$  \\
			$k$ & Number of selected elements from $Y_n$  \\
			$Y_n^k$ & Concatenated vector of $k$ selected elements from $Y_n$   \\
			$m$ & Number of multi-heads in self-attention $(>1)$ \\
			$h_{\text{head}}$ & hidden dimension for each head in multi-head self-attention \\
			$M_l^Q, M_l^K, M_l^V$ & Transform matrices of size $d \times h_{\text{head}}$ for $l$-th head in self-attention  $(1 \leq l \leq m)$ \\
			$M^O$ & Transform matrices of size $d \times d$ aggregating outputs of the heads in self-attention  \\
			$f$ & Row-wise softmax function in multi-head self-attention \\
			$W_l$ &  Weighting matrix $f \left( \frac{(B_n M_l^Q) (B_n M_l^K)^T }{\sqrt{h_{\text{head}}}} \right)$ $(1 \leq l \leq m)$ \\
			$A_l$ &  $f \left( \frac{(B M_l^Q) (B M_l^K)^T }{\sqrt{h_{\text{head}}}} \right) (B M_l^V)$ in self-attention \\
			$K_{\text{sp}}$ & Number of block latent variable samples used for model learning \\
			$D_n$ & $D_{KL} [ p_\theta ( b_n | B_n, B_{1:n-1} ) || q_\phi ( b_n | B_n, B_{1:n-1} ) ]$ \\
			$w_n^j $ & $ \frac{  p_\theta ( B_n, b_n^j |    B_{1:n-1}  )   }{ q_\phi ( b_n^j | B_n, B_{1:n-1} )  }$  \\
			$\pi$ & RL policy \\
			\bottomrule
		\end{tabular}
	\end{adjustbox}
	\caption{ Used notations in the main paper  }
	\label{tab:notations}
\end{table}

\newpage
\section{ Self-Attention Architecture } \label{sec:append_self_attention}

\begin{figure}[!ht] %%% Figure 1
	\begin{center} 
		\includegraphics[width=\linewidth]{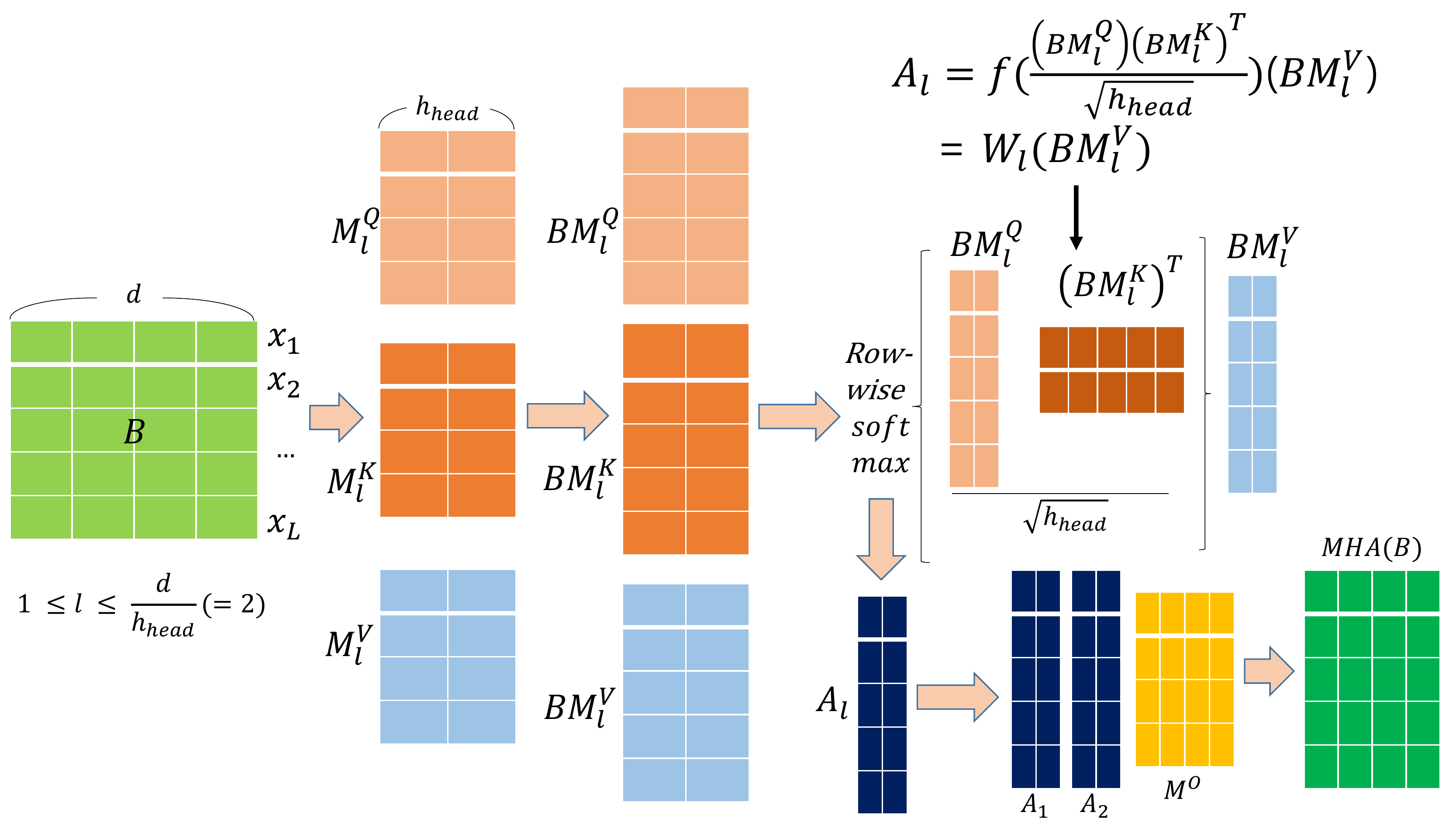}
% 		\vspace{-6mm}
		\caption{Self-attention architecture diagram describing the operation in Eq. \eqref{eq:multihead}  }
		\label{fig:attention_picture}
	\end{center}
% 	\vspace{-2mm}
\end{figure}

\newpage
\section{ Derivation of Gradient Estimation } \label{sec:append_grad_estimate}

\begin{figure}[!ht] %%% Figure 2
	\begin{center} 
		\includegraphics[width=0.6\linewidth]{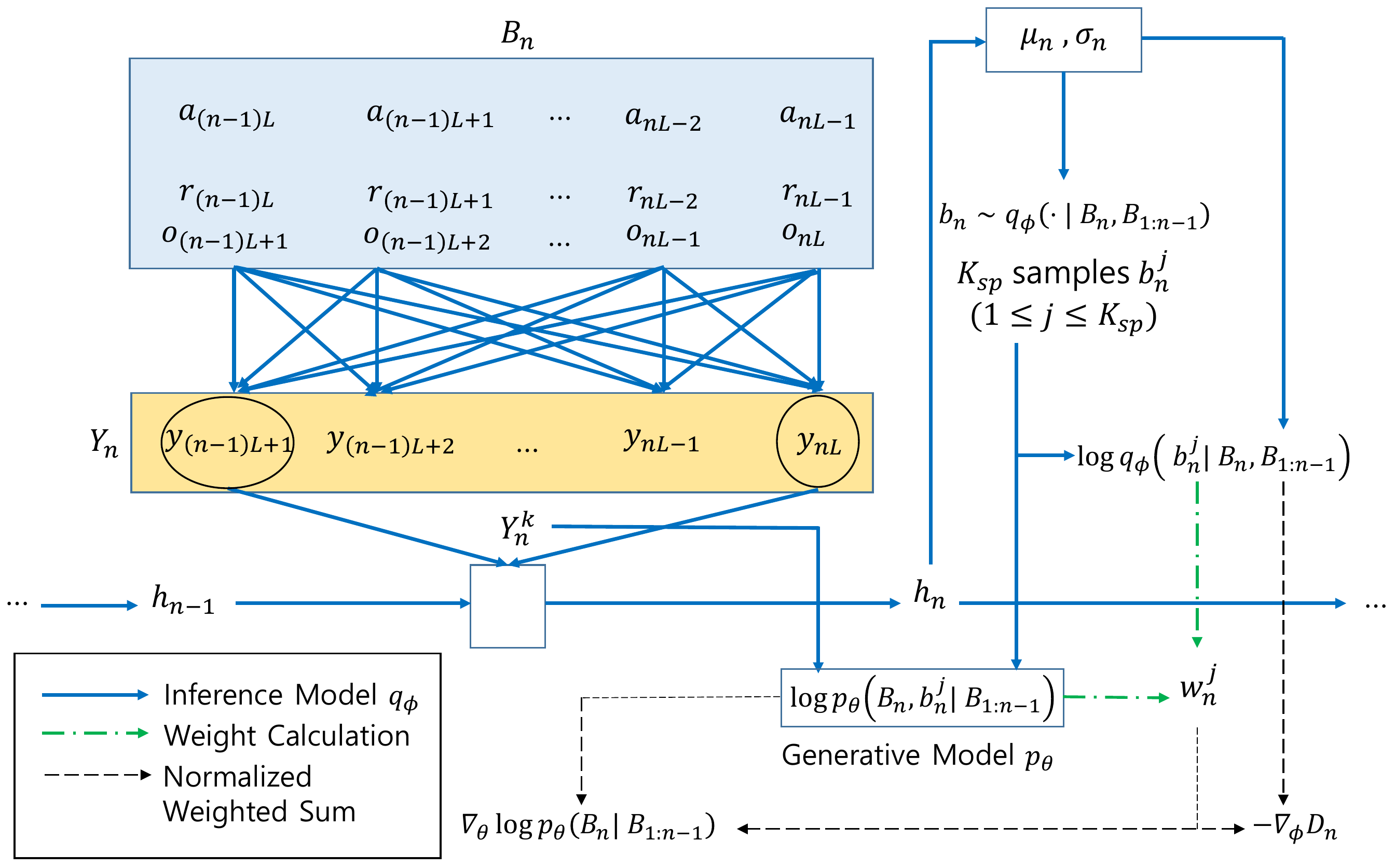}
		\caption{Estimation of $\nabla_{\theta} \log p_\theta ( B_n | B_{1:n-1} )$ and  $- \nabla_{\phi} D_n$}
		\label{fig:proposed_architecture_learning}
	\end{center}
\end{figure}

\begin{align} 
    \nabla_{\theta} \log p_\theta ( B_n | B_{1:n-1} )  &= \frac{ \nabla_{\theta}  p_\theta ( B_n | B_{1:n-1} )    }{  p_\theta ( B_n | B_{1:n-1} )   }   
    = \frac{ \nabla_{\theta} \int  p_\theta ( B_n, b_n | B_{1:n-1} ) d b_n   }{  \int  p_\theta ( B_n, b_n | B_{1:n-1} ) d b_n   }  \nonumber \\
    &= \frac{ \int \nabla_{\theta} p_\theta ( B_n, b_n | B_{1:n-1} ) d b_n   }{  \int  p_\theta ( B_n, b_n | B_{1:n-1} ) d b_n   }  \nonumber \\
    &=  \frac{ \int p_\theta ( B_n, b_n | B_{1:n-1} )  \nabla_{\theta} \log p_\theta ( B_n, b_n | B_{1:n-1} ) d b_n   }{  \int  p_\theta ( B_n, b_n | B_{1:n-1} ) d b_n   }  \nonumber \\
    &= \frac{  \int  q_\phi ( b_n | B_n, B_{1:n-1} ) \frac{ p_\theta ( B_n, b_n | B_{1:n-1} )  }{  q_\phi ( b_n | B_n, B_{1:n-1} )  } \nabla_{\theta} \log p_\theta ( B_n, b_n | B_{1:n-1} ) d b_n   }{  \int  q_\phi ( b_n | B_n, B_{1:n-1} ) \frac{ p_\theta ( B_n, b_n | B_{1:n-1} )  }{  q_\phi ( b_n | B_n, B_{1:n-1} )  }  d b_n     }  \nonumber \\
    & \approx \frac{ 1 }{ \sum_{j'=1}^{ K_{\text{sp}} } w_n^{j'} }  \sum_{j=1}^{ K_{\text{sp}} }  w_n^j \nabla_{\theta} \log p_\theta ( B_n, b_n^j |    B_{1:n-1}  ).
\end{align}

\begin{align} 
    - \nabla_{\phi} D_n &= - \nabla_{\phi} \int  p_\theta ( b_n | B_n, B_{1:n-1} ) \log \frac{p_\theta ( b_n | B_n, B_{1:n-1} )}{q_\phi ( b_n | B_n, B_{1:n-1} ) }  db_n  \nonumber\\
    &=  \int  p_\theta ( b_n | B_n, B_{1:n-1} )  \nabla_{\phi} \log q_\phi ( b_n | B_n, B_{1:n-1} )  d b_n      \nonumber \\
    &= \frac{ \int p_\theta ( B_n,  b_n | B_{1:n-1} )  \nabla_{\phi} \log q_\phi ( b_n | B_n, B_{1:n-1} )  d b_n    }{  \int p_\theta (  B_n, b_n | B_{1:n-1} ) db_n     }     \nonumber \\
    &= \frac{  \int  q_\phi ( b_n | B_n, B_{1:n-1} ) \frac{ p_\theta ( B_n, b_n | B_{1:n-1} )  }{  q_\phi ( b_n | B_n, B_{1:n-1} )  } \nabla_{\phi} \log q_\phi ( b_n | B_n, B_{1:n-1} )    d b_n   }{  \int  q_\phi ( b_n | B_n, B_{1:n-1} ) \frac{ p_\theta ( B_n, b_n | B_{1:n-1} )  }{  q_\phi ( b_n | B_n, B_{1:n-1} )  }  d b_n     }  \nonumber \\
    % &= \mathbb{E}_{ p_\theta ( b_n | B_n, B_{1:n-1} )  }  [ \nabla_{\phi} { K_{\te}  w_n^j \nabla_{\phi} \log q_\phi ( b_n^j | B_n, B_{1:n-1} ).
    &\approx \frac{ 1 }{ \sum_{j'=1}^{ K_{\text{sp}} } w_n^{j'} } \sum_{j=1}^{ K_{\text{sp}} }  w_n^j \nabla_{\phi} \log q_\phi (  b_n^j | B_n,   B_{1:n-1}  ).
\end{align}

\newpage
\section{ Algorithm } \label{sec:append_algorithm}

We have two versions of the proposed method: Algorithm \ref{algorithm:main_algorithm_off_policy} for off-policy RL learning and Algorithm \ref{algorithm:main_algorithm_on_policy} for on-policy method, respectively. 

\begin{algorithm}[!ht] 
	\caption{Blockwise Sequential Model Learning with Off-Policy RL Update}\label{algorithm:main_algorithm_off_policy}
	\begin{algorithmic}[1]
		\STATE{ $L:$ Block length, $T$: Sample sequence length ($T=NL$ with the number of blocks $N$ ) }
		\STATE{ $k:$ Number of selected elements from $Y_n$, $MAX:$ Maximum timestep for training }
		\STATE{$I_{\text{RL}}: $ RL training interval, $I_{\text{model}}: $  Block model learning interval}
		\STATE{$I_{\text{pre}}: $ Timestep before pre-training the block model, $S_{\text{pre}}: $  Number of pre-training steps}
		\STATE{$\mathcal{M}$: Replay memory,  $N_{\text{mini}}$: Minibatch size for learning, $\text{SG}$: Stop gradient    }
		\STATE{$t_{\text{global}}$: global timestep, $t$: time index within a block, $n:$ block index    }
		%%%% Begin
		\STATE{Set $ t_{\text{global}} = 0 $.    }
		\WHILE{Timestep $ t_{\text{global}} < MAX$}
		    \STATE{ Initialize a new episode $E$ with initial observation $o_0$ from the environment.  }
		    \STATE{ Initialize the first latent variable of RNN in the block model as $h_0 = \boldsymbol{0}$, and generate $\mu_0$ and $\sigma_0$ in the model.  }
		    \STATE{ Initialize the RL input variable $z_0 = \boldsymbol{0}$.  }
		    \STATE{ Set $t = 0$, $n=1$.  }
    		\WHILE{ $E$ is not ended    }
    		    \STATE{ Select action $a_t$ from $\pi(\cdot | z_t, o_t)$ and receive $(r_{t}, o_{t+1})$ from the environment. Save $x_{t+1} = (a_{t}, r_{t}, o_{t+1})$ in $\mathcal{M}$.  }
    		    \STATE{ Generate $z_{t+1}$ from the RNN for RL input given the inputs of $x_{t+1}$ and $\text{SG}(\mu_{n-1}, \sigma_{n-1})$.   } 
        		\IF{$ t  \% L == L-1$ } 
            		\STATE{ Generate $Y_n$ of the output of the self-attention given the input block $B_n = x_{ t - L + 2 : t + 1 }$.	}
            		\STATE{ Select $k$ elements $Y_n^k$ among $Y_n$.	}
            		\STATE{ Generate the model RNN output $h_n$ with the input  $Y_n^k$. Generate $\mu_n$ and $\sigma_n$ using $h_n$. }
            		\STATE{ $n \leftarrow n+1$. }
            	\ENDIF
            	\STATE{ $t \leftarrow t+1, t_{\text{global}} \leftarrow t_{\text{global}}+1$.   }
            	\IF{$ t_{\text{global}} == I_{\text{pre}} + 1$ } 
            	    \STATE{ Sample $N_{\text{mini}}$ number of sequences $B_{1:N}$ from $\mathcal{M}$ and update $q_\phi$ and $p_\theta$ using \eqref{eq:generative_learn} and \eqref{eq:inference_learn}, respectively for $S_{\text{pre}}$ times.  } \\
            	    \hfill\COMMENT{// Pre-training $q_\phi$ and $p_\theta$} 
            	\ENDIF
            	\IF{$ t_{\text{global}} > I_{\text{pre}}$ and $t_{\text{global}} \% I_{\text{RL}} == 0$}
            	    \STATE{ Sample $N_{\text{mini}}$ number of sequences $B_{1:N}$ from $\mathcal{M}$ and generate $Z_{1:N}$ from the RNN for RL input. Update the off-policy RL agent by using $Z_{1:N}$ and $B_{1:N}$ as inputs. } \hfill\COMMENT{// RL update} 
            	\ENDIF
            	\IF{$ t_{\text{global}} > I_{\text{pre}}$ and $t_{\text{global}} \% I_{\text{model}} == 0$}
            	    \STATE{ Sample $N_{\text{mini}}$ number of sequences $B_{1:N}$ from $\mathcal{M}$ and update $q_\phi$ and $p_\theta$ using \eqref{eq:generative_learn} and \eqref{eq:inference_learn}, respectively for once. } \\
            	    \hfill\COMMENT{// Learning $q_\phi$ and $p_\theta$} 
            	\ENDIF
        	\ENDWHILE
		\ENDWHILE
	\end{algorithmic}
\end{algorithm}

\newpage
%% Add on policy version as well.

\begin{algorithm}[!ht] 
	\caption{Blockwise Sequential Model Learning with On-Policy RL Update}\label{algorithm:main_algorithm_on_policy}
	\begin{algorithmic}[1]
		\STATE{ $L:$ Block length, $k:$ Number of selected elements from $Y_n$}
		\STATE{ $MAX:$ Maximum observation number for training }
		\STATE{$I_{\text{update}}: $ Interval for block model learning and RL training}
		\STATE{ $\mathcal{B}$: Current block batch, $\text{SG}$: Stop gradient    }
		\STATE{$t_{\text{global}}$: global timestep, $t$: time index within a block, $n:$ block index    }
		%%%% Begin
		\STATE{Set $ t_{\text{global}} = 0 $.    }
		\WHILE{Timestep $ t_{\text{global}} < MAX$}
		    \STATE{ Initialize a new episode $E$ with initial observation $o_0$ from the environment.  }
		    \STATE{ Initialize the first latent variable of RNN in the block model as $h_0 = \boldsymbol{0}$, and generate $\mu_0$ and $\sigma_0$ in the model.  }
		    \STATE{ Initialize the RL input variable $z_0 = \boldsymbol{0}$.  }
		    \STATE{ Set $t = 0$, $n=1$.  }
    		\WHILE{ $E$ is not ended    }
    		    \STATE{ Generate $z_{t+1}$ from the RNN for RL input given the inputs of $o_t$ and $\text{SG}(\mu_{n-1}, \sigma_{n-1})$.   } 
    		    \STATE{ Select action $a_{t+1}$ from $\pi(\cdot | z_{t+1})$ and receive $(r_{t+1}, o_{t+1})$ from the environment.   }
        		\IF{$ t  \% L == L-1$ } 
            		\STATE{ Generate $Y_n$ of the output of the self-attention given the input block $B_n = o_{ t - L + 1 : t  }$.	}
            		\STATE{Save $B_n$ in the current block batch $\mathcal{B}$. }
            		\STATE{ Select $k$ elements $Y_n^k$ among $Y_n$.	}
            		\STATE{ Generate the model RNN output $h_n$ with the input  $Y_n^k$. Generate $\mu_n$ and $\sigma_n$ using $h_n$. }
            		\STATE{ $n \leftarrow n+1$. }
            	\ENDIF
            	\STATE{ $t \leftarrow t+1, t_{\text{global}} \leftarrow t_{\text{global}}+1$.   }
            % 	\IF{$ t_{\text{global}} == I_{\text{pre}} + 1$ } 
            % 	    \STATE{ Sample $N_{\text{mini}}$ number of sequences $B_{1:N}$ from $\mathcal{M}$ and update $q_\phi$ and $p_\theta$ using \eqref{eq:generative_learn} and \eqref{eq:inference_learn}, respectively for $S_{\text{pre}}$ times.  } \hfill\COMMENT{// Pre-training $q_\phi$ and $p_\theta$} 
            % 	\ENDIF
                \IF{ $t_{\text{global}} \% I_{\text{update}} == 0$}
            	    \STATE{ Use the current block batch $\mathcal{B}$ to update $q_\phi$ and $p_\theta$ using \eqref{eq:generative_learn} and \eqref{eq:inference_learn}, respectively. } \hfill\COMMENT{// Learning $q_\phi$ and $p_\theta$} 
            	    \STATE{ Generate the RL input $Z$ from the RNN for RL input with $\mathcal{B}$ and updated $q_\phi$.}
            	    \STATE{ Update the on-policy RL agent using $Z$. } 
            	    \hfill\COMMENT{// RL update} 
            	    \STATE{ Empty $\mathcal{B}$. }
            	\ENDIF
        	\ENDWHILE
		\ENDWHILE
	\end{algorithmic}
\end{algorithm}

\newpage
\section{ Details in Implementation } \label{sec:append_implementation}

We describe implementations with (i) off-policy RL agent in Mountain Hike, Pendulum, sequential target-reaching task, and (ii) on-policy RL agent in Minigrid.

\subsubsection{Off-policy Implementation}

For the actual implementation of the proposed method and the performance comparison, we modified the open-source code of VRM \cite{han20vrm}, which includes the sequential target-reaching environment and the codes of LSTM and SAC. We implemented the source code of the proposed method using Pytorch, and we used Intel Core i7-7700 CPU 3.60GHz and i7-8700 CPU 3.20GHz as our computing resources. We focused on the performance comparison within a fixed sample size rather than the average runtime or estimated energy cost. We used the code of Mountain Hike environment from \cite{igl18dvrl}, and we modified the POMDP wrapper code provided by \cite{meng21memory} for the considered Pendulum environment. The episode length is 200 timesteps in the considered Pendulum, and the maximum episode length is 128 timesteps in the sequential target-reaching task.

In the original code of VRM, $\tanh$ activation is added at the end of the policy for stable action selection. We used this technique in the proposed method, LSTM, and SAC for fair comparisons. We also checked that VRM with explicit gradient calculation of the actor loss in the RL agent learning performed better than directly applying Adam optimizer to the actor loss. Therefore, we used the explicit gradient calculation method in the proposed method, LSTM, and SAC for fair comparisons. The proposed method, VRM, and LSTM sample data sequence from the replay memory $\mathcal{M}$ of length $T=64$ with minibatch size $N_{\text{mini} } = 4$. For the implementation of RIMs, we used the open-source code \cite{goyal21rims} as a drop-in replacement of LSTM. We set the dimension of the hidden variable of RIMs to be $384 = 64 \times 6$, and we tuned the number of selected segments $k_{\text{RIMs}} < 6$ to produce the best performance for each three environments. We found the best $k_{\text{RIMs}} $ as 3 in Mountain Hike and the sequential target-reaching task, and 4 for the considered Pendulum.

For the inference model $q_{\phi}$, we first convert $x_t = [a_{t-1}; r_{t-1}; o_t]$ into $ \tilde{x}_t =   [\varphi_{\text{model}}( o_t; r_{t-1} );  a_{t-1} ] \in \mathbb{R}^{256} $,  where $\varphi_{\text{model}}$ is a feed-forward neural network with the hidden layer size 256 and $\tanh$ activation. Then, the $n$-th block $B_n = \tilde{x}_{(n-1)L+1:nL}$ is fed into the self-attention network without positional encoding. The number of multi-heads is set to $m=4$, so $h_{\text{head}}= 256 / 4 = 64$. The conventional dropout in self-attention after $\text{MHA}(B)$ and $g(U)$ in \eqref{eq:attention_all} is used with dropout rate 0.1, and dropout is turned off during action selections from the RL policy. $g(\cdot)$ in \eqref{eq:attention_all} is a feed-forward neural network with the hidden layer size 512 and Gaussian error linear units (GELU) activation. We observed that a stack of more than two self-attentions gave a meaningful difference among the elements of the column-wise summation of $\frac{1}{m}\sum_{l=1}^m W_l^{\text{last}}$, where $W_l^{\text{last}}$ is the weighting matrix at the last self-attention network. We stack two self-attentions for the block model and selected $Y_n^k \in \mathbb{R}^{256 k}$ from the output $Y_n$ of the two-stacked  self-attentions. The values of $L$ and $k$ were selected among $L \in \{ \frac{T}{2^i} \}_{i=1}^5$ and $k$ satisfying $k \leq L/4$. The proposed method uses $L=16, k=2$ for Mountain Hike, $L=32, k=2$ for the considered Pendulum, and $L=32, k=3$ for the sequential target-reaching task. (Therefore, in Section \ref{subsec:ablation_components_compare}, we set $S_{\text{FNN}} = 24, 16$ for the sequential target-reaching task with $R=15$ and the considered Pendulum, respectively. We used MLP with two hidden layers and $\tanh$ activation for FNN.)  When a sample sequence has  length less than $k$, zero padding is used to meet the dimension of $Y_n^k \in \mathbb{R}^{256 k}$.

The RNN for the block model is a GRU \cite{cho14gru} with the hidden variable $h_n \in \mathbb{R}^{256}$ at the $n$-th block. $\mu_n \in \mathbb{R}^{64}$ is the output of a neural network with the hidden layer size 128 and $\tanh$ activation. $\sigma_n \in \mathbb{R}^{64}$ is the output of another neural network with the hidden layer size 128 and $\tanh$ activation, and the output is followed by softplus activation. $b_n^j \in \mathbb{R}^{64} (1 \leq j \leq K_{\text{sp}})$ is sampled for $K_{\text{sp}}=50$ times for the block model learning. For the generative model $p_{\theta}$, the concatenation $Y_n^k$ and each $b_n^j  $ is fed into the constructed generative model network with the hidden layer size 256 and $\tanh$ activation to produce $\log p_\theta(B_n, b_n^j | B_{1:n-1}) $.

For the RL optimization,  we convert $x_t = [a_{t-1}; r_{t-1}; o_t]$ at the $n$-th block into $ \hat{x}_t =   [\varphi_{\text{model}}( o_t; r_{t-1} );  a_{t-1} ] \in \mathbb{R}^{256} $, where $\varphi_{\text{model}}$ is a feed-forward neural network with the hidden layer size 256 and $\tanh$ activation. Note that $\varphi_{\text{model}}$ is learned with block model update. Then, $[\hat{x}_t; \text{SG}(\mu_{n-1}; \sigma_{n-1})]$ is fed into the RL input generation RNN which is another GRU with the hidden variable $z_t \in \mathbb{R}^{256}$ (SG refers to stopping the gradient). The policy, value function, target value function, and two Q-functions are constructed using feed-forward neural networks, each of which has two hidden layers of size $(256, 256)$ and ReLU activations. At each timestep $t$, $[z_t; o_t; r_{t-1}]$ is fed into the policy network and the value function network, and $[z_t; o_t; r_{t-1}; a_{t-1}]$ is fed into the Q-function network. We pretrain the block model $S_{\text{pre}}=500$ times after $I_{\text{pre}}=1000$ timesteps before RL learning begins. Then, we periodically train both the block model and the  RL agent with $I_{\text{RL}}=1$ and $I_{\text{model}}=5$ timesteps, respectively.  Adam optimizer is used for training, and the learning rates of the block model and the  RL agent are $8e-4$ and $3e-4$, respectively.

\subsubsection{ On-policy Implementation }

For the actual implementation of the proposed method and the performance comparison, we modified the open-source code of `rl-starter-files' in github which includes the PPO and LSTM implementations in the Minigrid environment. In this code, every $7 \times 7 \times 3$ observation $o_t$ is encoded by a convolutional neural network $\varphi_{\text{RL}}$ to produce $\varphi_{\text{RL}}(o_t) \in \mathbb{R}^{64}$. Note that $\varphi_{\text{RL}}$ is learned with RL update. We implemented the source code of the proposed method using Pytorch, and we used TITAN Xp GPU and GeForce RTX 2060 GPU as our computing resources. We focused on the performance comparison within a fixed sample size rather than the average runtime or estimated energy cost. 
As above, for the implementation of RIMs, we used the open-source code \cite{goyal21rims} as a drop-in replacement of LSTM. We set the dimension of the hidden variable of RIMs to be $192 = 32 \times 6$. We found that using 192 instead of 64 (which is the dimension of $\varphi_{\text{RL}}(o_t)$ ) performed better. We tuned the number of selected segments $k_{\text{RIMs}} < 6$ to produce the best performance for each environment. We found the best $k_{\text{RIMs}} $ as 5 in CrossingS9N3 and CrossingS11N5.

The block model is the same as above except the dimension of $Y_n^k$ is $64k$ and there is no pretraining. The values of $L$ and $k$ were selected among $L \in \{ 4, 8, 16, 32, 64 \}$ and $k$ even number satisfying $k \leq L$. The proposed method uses $L=8, k=4$ for CrossingS9N3, and $L=16, k=14$ for CrossingS11N5. 16 same environments are run in parallel for actual implementation, so we set $I_{\text{update}} = 128 \times 16 = 2048$. (See Appendix \ref{subsec:append_minigrid} for the details.) The block model is updated with 2 epochs given the current batch of size $I_{\text{update}} = 2048$. The value of $\log \sigma_n$ is clipped between -20 and 2, and the overall gradient norm is clipped to 0.1.
For RL update,  $[\varphi_{\text{RL}}(o_t); \text{SG}(\mu_{n-1}; \sigma_{n-1})] \in \mathbb{R}^{192}$ is fed into the RL input generation RNN which is LSTM with the hidden variable $z_t \in \mathbb{R}^{64}$ (SG refers to stopping the gradient). The policy and value function are constructed using feed-forward neural networks, each of which has a hidden layer of size $(64, 64)$ and $\tanh$ activations.  Adam optimizer is used for training, and the learning rates of the block model and the  RL agent are $1e-5$ and $1e-3$, respectively.

%%%% 

% \newpage
\section{ Details in Environments } \label{sec:append_environment}

\subsection{ Mountain Hike }

State $s_t$ is a two-dimensional position of the agent, and observation $o_t$ is given by $o_t = s_t + \epsilon_{\text{error},t}$, where $\epsilon_{\text{error},t} \sim \mathcal{N}(0, \sigma_{\text{error}}^2 I)$. The inital state is given by $s_0 \sim \mathcal{N}((-8.5,-8.5), I)$ for every episode. Action is $a_t = [d x_t, d y_t]$ and the agent receives next state $s_{t+1} = s_t + \hat{a}_t + \epsilon_{\text{trans}, t}$, where $\hat{a}_t$ is the normalized vector of $a_t $ satisfying $ \| \hat{a}_t \| = \min \{ c_{\text{thres}}, \| a_t \| \}  $ and $\epsilon_{\text{trans}, t} \sim \mathcal{N}(0, 0.25  I)$. The agent receives reward $r_t = r_t^{\text{map}}(x_t, y_t) - 0.01\| a_t \|$, where $r_t^{\text{map}}(x_t, y_t)$ is shown in Fig. \ref{fig:MTHike_env_map}. To enforce the larger partial observability than original environment of \citet{igl18dvrl}, we set the noise variance $\sigma_{\text{error}}^2 = 3.0$ and reduced the maximum action norm $c_{\text{thres}}=0.1$ from the original value 0.5 in \citet{igl18dvrl}. We instead increased the episode length from 75 to 200 to observe the performance improvement more accurately.

\subsection{Minigrid} \label{subsec:append_minigrid}

For the considered two environments CrossingS9N3 and CrossingS11N5 in Minigrid, S means the map size, and N is the maximum crossing paths. For example, in CrossingS9N3, a new map with the same $9 \times 9$ size but different shapes with three crossing paths at most is generated every episode. Then an agent (red triangle in Figs. \ref{fig:minigrid_S9N3_map}, \ref{fig:minigrid_S11N5_map}) starts at the upper-left corner of each map to find the location of the green square. The agent has $7 \times 7$ egocentric partial observation with each tile encoded as a three-dimensional input, including information on objects, color, and state. Therefore, the total input dimension at each timestep is $7 \times 7 \times 3$. The agent can choose actions such as `move forward,' `turn left,' 'turn right.' For CrossingS9N3, the episode ends if the agent succeeds in reaching the goal or $N_{\text{max}}=324$ timesteps passed. For CrossingS11N5, we have $N_{\text{max}}=484$. For the actual implementation based on PPO \cite{schulman17ppo}, 16 same environments are run in parallel.

% \newpage
\section{Effect of Hyperparameters } \label{sec:append_ablation_hyperparameter}

We experimented by changing the hyperparameter $L$ (block length) and $k$ (number of selected elements in the self-attention output) in the sequential target-reaching task with $R=15$. The success rates of the proposed method with $L=4, 8, 16, 32$ given the same $k(=3)$ are $32.0\%$, $19.6\%$, $57.8\%$, $\mathbf{91.4\%}$, respectively. As $L$ becomes small, the proposed block model approaches the stepwise recurrent neural network. $L=4, 8$ have difficulty passing the relative blockwise information to the following blocks.

The success rates of the proposed method with $k=1, 3, 5, 7$ given the same $L(=32)$ are $67.7\%$, $\mathbf{91.4\%}$, $67.3\%$, $66.6\%$, respectively. Recall that we calculated the $\log p_\theta(B_n, b_n^j | B_{1:n-1} )$, where $b_n^j \sim q_\phi( \cdot | B_n, B_{1:n-1})$ (see Fig. \ref{fig:proposed_architecture_learning} in Appendix \ref{sec:append_grad_estimate}). As $k$ increases, the dimension of the self-attention output $Y_n^k (=256 k)$ becomes larger.  Then the portion of block variable $b_n^j$ in $\log p_\theta(B_n, b_n^j | B_{1:n-1} )$ becomes smaller than the portion of $B_n$. In this case, the proposed self-normalized importance sampling may not work properly.  Therefore, the performance degrades with $k$ larger than some optimal threshold values ($k=3$ in this case).

% \newpage
\section{Related Work on Sequential Representation Modeling }
Several works use graphical models to learn latent variables given sequential data inputs. \citet{chung15vrnn} used a generative model  and an inference model to maximize the variational lower bound of $\log p_{\theta}(x_{1:T})$. The past data information is compressed using RNN in a stepwise manner to estimate the latent variables. \citet{li18seqae} proposed two types of encoder for sequential latent variable modeling: unconditional and conditional generation. Recently, \citet{han2021disentangled} proposed a robust sequential model learning method using  a disentangled (unconditional) structure based on the Wasserstein metric \cite{arjovsky17wgan}. Unlike these sequential learning methods using decoder networks to reconstruct the input data, the proposed method does not require sequential data reconstruction.  In addition, the proposed model is trained in a blockwise manner, unlike other methods considering stepwise learning approaches.

% \newpage
\section{ Limitation and Future Work }

We used the self-attention network to produce the output $Y_n$ capturing the contextual information in the $n$-th block $B_n$, and we selected the representative $k$ elements $Y_n^k$ in $Y_n$  with largest $k$ contributions in column-wise summation of $\frac{1}{m} \sum_{l=1}^m W_l$ in timestep. Although we consider the one-to-one mapping from $B_n$ to $Y_n$, the column-wise summation of $W_l$ reflects the overall contribution of each row  of  $B_n M_l^V$, not $A_l$ or $Y_n$. This method implies a discrepancy between the column-wise summation of the weighting matrices and the importance of $Y_n^k$, even though the proposed approach performs well in practice as seen in Section \ref{sec:experiments} and \ref{sec:ablation_study}. Our future work will theoretically investigate which conditions are required for some compression methods to guarantee the optimality of learning procedures. Another future direction would be designing an adaptive scheduler of the value of $k$ to deal with information redundancy and hyperparameter sensitivity.

\section{ Social Impact }

This research is about RL in partially observable environments. Partially observable RL occurs in many real-world control problems where total observations of the actual underlying states are not available. For instance, an uncrewed aerial vehicle (UAV) such as a drone should control its movement and navigate to the target position even when the current position information is distorted due to the noise or malfunction of the position sensor. In this case, past information should be exploited to estimate the actual position. Advances in research on partially observable RL will improve many automated control tasks. We expect our work can benefit society and industry by improving robotic control accuracy or reducing factory operation malfunction. On the other hand, too high efficiency on industrial automation may cause job loss in some regions of society. However, this problem should be solved through social security plan based on the wealth generated by new technologies.

\end{document}